\documentclass[a4paper,fleqn]{cas-sc}

\usepackage[numbers,sort&compress]{natbib}

\def\tsc#1{\csdef{#1}{\textsc{\lowercase{#1}}\xspace}}
\tsc{WGM}
\tsc{QE}

\begin{document}
\let\WriteBookmarks\relax
\def\floatpagepagefraction{1}
\def\textpagefraction{.001}

\shorttitle{ADAPT-CAM}    

\shortauthors{Sahar et al.}  

\title [mode = title]{One Size Doesn’t Fit All: Input Adaptive Class Activation Mapping for Improved Visual Explanations}

%
\author[1]{Ayesha~Sahar}
\cormark[1]
\fnmark[1]
\ead{ayeshasahar020@gmail.com}
\credit{Writing -- original draft, Conceptualization, Formal analysis, Software, Methodology, Visualization}

\affiliation[1]{organization={Department of Computer Science, Air University},
            country={Pakistan}}

\author[2]{Tahir~Mustafa~Madni}
\fnmark[2]
\ead{tahir_mustafa@comsats.edu.pk}
\credit{Writing -- review \& editing, Supervision}

\author[2]{Tehseen~Zia}
\fnmark[3]
\ead{tehseen.zia@comsats.edu.pk}
\credit{Writing -- review \& editing, Supervision}

\author[3]{David~Windridge}
\fnmark[4]
\ead{D.Windridge@mdx.ac.uk}
\credit{Writing -- review \& editing}

\author[2]{Uzair~Iqbal~Janjua}
\fnmark[5]
\credit{Writing -- review \& editing}

\author[2]{Waleed~Umar}
\fnmark[6]
\ead{waleedumar85@gmail.com}
\credit{Writing -- original draft, Validation, Visualization, Data Curation}

\affiliation[2]{organization={Department of Computer Science, COMSATS University Islamabad},
            city={Islamabad},
            country={Pakistan}}

\affiliation[3]{organization={Department of Computing, Middlesex University},
            city={London},
            country={U.K.}}

\cortext[1]{Corresponding author}

\begin{abstract}
Despite their widespread use in tasks such as image classification, object detection and semantic segmentation, the limited interpretability of Convolutional Neural Networks (CNNs) and Vision Transformers (ViTs) can undermine trust and hinder adoption in safety-critical applications. Explainable artificial intelligence (XAI) methods seek to address this by providing human-understandable explanations for model predictions. Class Activation Maps (CAM) constitute one such widely used XAI technique for visualizing discriminative regions. However, existing CAM methods may exhibit limitations in channel weighting, localization, and adaptability across network architectures potentially resulting in noisy or imprecise explanations. To address these issues, we propose ADAPT-CAM, a gradient-based, parameter-adaptive CAM framework. The method combines first-order gradient energy with activation-weighted second-order curvature to derive channel contributions. It further incorporates adaptive local contrast normalization and knee-guided soft channel selection to refine the resulting explanations. Importantly, key parameters are estimated independently for each input, allowing the framework to adapt to variations in feature representations across network architectures. We validate ADAPT-CAM on ImageNet and Oxford 102 Flowers across CNN and ViT architectures. Quantitative results demonstrate competitive or superior performance against established CAM methods across multiple evaluation metrics. In parallel, qualitative comparisons illustrate its ability to highlight discriminative image regions. Further parameter analysis and ablation experiments examine the behaviour and contributions of the individual adaptive components. These findings demonstrate the applicability of ADAPT-CAM across different architectures and datasets.
\end{abstract}



\maketitle

\section{Introduction}
Deep convolutional neural networks (CNNs) have achieved remarkable performance in computer vision tasks, including image classification, object detection, semantic segmentation, and fine-grained visual recognition \cite{b5, b6, b7, b8, b9}. Despite their success, their black-box nature limits transparency and raises concerns about their reliability in safety-critical applications \cite{b27, b28, b29}. For instance, in medical imaging, a CNN may identify a tumour without revealing whether its prediction relies on clinically meaningful features or irrelevant imaging artifacts \cite{b4, b24}. Similarly, autonomous driving systems may detect pedestrians or obstacles without providing an interpretable explanation of the visual evidence underlying their predictions \cite{b25}. Such limitations highlight the need for explainable artificial intelligence (XAI), which aims to make model decisions more transparent, interpretable and accountable \cite{b26, b30}.

XAI methods can broadly be classified into ante-hoc and post-hoc approaches. While ante-hoc methods incorporate interpretability into the model architecture, post-hoc methods explain predictions after training without requiring modifications to the underlying model \cite{b26, b22}. This study focuses on post-hoc approaches, particularly Class Activation Mapping (CAM) \cite{b11} and its derivatives, which generate spatial explanations by identifying image regions relevant to a target-class prediction. The original CAM formulation uses class-specific weights to combine convolutional feature maps but requires a particular network architecture. Subsequent CAM methods introduced gradient-based and gradient-free approaches to improve the localization and class-discriminative quality of activation maps. In addition, several hybrid and fusion-based approaches have been developed to enhance explanation consistency as well.

Despite this progress, existing CAM methods retain several limitations. First, channel importance is typically estimated using spatially aggregated gradient statistics. As a result, the spatial distribution of gradient sensitivity and local contrast within each feature map does not directly inform how channels are scored. Consequently, channels responding to class-relevant local structures are often not explicitly distinguished from those showing diffuse activations. Second, many methods lack an explicit adaptive channel selection mechanism and allow channels with relatively weak contributions to influence the final explanation. Finally, many channel-importance and selection strategies rely on parameters that are typically fixed constants shared across all inputs. Therefore, they cannot adapt to differences in feature distributions, texture or discriminative granularity across images and domains.

Biological vision offers useful insights into addressing these limitations. In the early visual system, center-surround interactions and divisive normalization enable neurons to process visual information relative to its surrounding context which enhances local contrast and regulates neuronal responses \cite{b45}. Furthermore, competitive interactions and attentional gain control in visual processing modulate the relative responses of neuronal populations. This allows relevant information to receive greater emphasis while less relevant responses are attenuated rather than entirely suppressed \cite{b46}. These biological processes highlight the importance of contextual processing, adaptive response regulation and selective information representation in visual perception.

Taking inspiration from these principles, we propose ADAPT-CAM, a gradient-based, parameter-adaptive CAM framework. It first combines first-order gradient energy with an activation-weighted second-order curvature term to produce power-scaled channel weighting.  Motivated by contextual contrast processing, we incorporate adaptive Local Contrast Normalization (LCN). Its neighborhood size is determined through a participation-ratio-based mechanism to adjust spatial contrast evaluation according to the distribution of gradient-weighted activations. Drawing inspiration from selective response modulation, a cumulative contribution-based knee-point detection mechanism identifies the transition between dominant and lower-ranked channels. It is followed by a knee-guided exponential rank-based mask that progressively attenuates lower-ranked channels without completely discarding them. The key parameters governing these operations are estimated independently for each input rather than fixed across the dataset or model architecture. This reflects the stimulus-dependent nature of biological visual processing in which neuronal responses adapt to the specific characteristics of each visual stimulus rather than remaining uniform across different inputs. Together, these complementary components form a balanced design that produces precise, stable and class-specific activation maps.

The main contributions of this work are summarized as follows:

\begin{itemize}

\item ADAPT-CAM is introduced as a unified formulation whose key parameters are derived per image allowing adaptation to different visual domains (e.g. medical vs. natural images) and task granularities e.g. coarse, fine-grained, weak-signal).

\item A power-scaled, curvature-modulated channel weighting formulation is developed by combining first-order gradient energy with activation-weighted second-order information to refine channel importance.

\item Adaptive LCN is incorporated with its neighborhood size determined through the participation ratio of gradient-weighted activations to enhance local contrast and spatial detail.

\item An adaptive soft channel selection mechanism is introduced using cumulative contribution-based knee-point detection followed by a knee-guided exponential rank-based mask that progressively attenuates lower-ranked channels to identify the dominant contribution transition.

\item The proposed framework is quantitatively evaluated against established CAM baselines across CNN and Transformer architectures to assess its localization performance.

\end{itemize}

\section{Related Work}
As deep networks grow more complex, XAI methods have become essential for understanding model decisions. Among visualization approaches, CAM and its variants are the dominant family of saliency-based methods. This section reviews the development of CAM-based methods, their underlying approaches and the limitations motivating our proposed framework.

\subsection{Gradient-Based CAM Methods}
The original CAM \cite{b11} weights the final convolutional feature maps with the fully connected weights that follow global average pooling (GAP), which restricts it to specially designed or re-trained CNNs. Grad-CAM \cite{b12} overcomes this by computing channel weights from spatially averaged first-order gradients of the class score, making it applicable to any differentiable network and a standard baseline. However, this averaging yields coarse maps that miss small or low-contrast features. To overcome this issue, Grad-CAM++ \cite{b13} incorporates higher-order derivative terms into the weighting which improves localization. However, its fixed formulation remains sensitive to gradient noise. Smooth Grad-CAM++ \cite{b14} then reduces this noise by averaging maps over noisy copies of the input, though this can suppress fine discriminative detail. XGrad-CAM \cite{b18} instead derives weights from activations and gradients under sensitivity and conservation axioms, yet it still depends on gradient reliability in deeper layers. More recently, methods such as  More recently, methods such as SSG-CAM \cite{b40} smooths the gradients with a Gaussian kernel and adds a second-order term to ease gradient noise and saturation.

\subsection{Gradient-Free CAM Methods}
To avoid any gradient based issues that affects these methods, Score-CAM \cite{b15} weights each activation map by the change in target-class confidence when the input is masked with it. This is more stable and gives sharper maps, but it depends heavily on the model's confidence distribution and may emphasize highly activated yet class-irrelevant regions. Eigen-CAM \cite{b16} then tries to remove the dependence on both gradients and confidence by projecting the feature maps onto their principal component obtained via SVD, making it fast and fully gradient-free. However, its class-agnostic nature can highlight globally dominant structures rather than class-discriminative ones.

\subsection{Multi-Layer and Fusion-Based CAM Methods}
To recover the spatial detail that single-layer maps lose, Layer-CAM \cite{b17} aggregates pixel-level gradients across several layers. It improves spatial precision and boundary delineation, but this aggregation may amplify non-target activations. Master-CAM \cite{b19} improves on this by fusing CAMs from large-scale inputs through a Master map, which strengthens contextual localization. However, noise in the Master map can propagate and weak features may be suppressed. Ensemble-CAM \cite{b20} instead combines several CAM variants for more consistent explanations, yet averaging reduces contrast and blurs boundaries. To restore this sharpness, Augmented Grad-CAM++ \cite{b23} adds augmentation and super-resolution, but it can amplify irrelevant noise and leaves gradient-level issues unresolved. Grad++Score-CAM \cite{b21} then tackles these gradient issues by combining Grad-CAM++ \cite{b13} sensitivity with Score-CAM \cite{b15} confidence weighting. Still, fusing the two signals may amplify irrelevant regions and produce diffuse boundaries in complex scenes.

\subsection{Motivation for Proposed Framework}
Despite their differences, existing CAM methods share several limitations. First, channel importance is typically estimated using spatially aggregated gradient statistics. Therefore, the spatial distribution of gradient sensitivity and local contrast within each feature map does not directly inform how channels are scored. Consequently, channels responding to class-relevant local structures are often not explicitly distinguished from those showing diffuse activations. Second, many methods lack an explicit adaptive channel selection mechanism and allow channels with relatively weak contributions to influence the final explanation. Finally, many channel-importance and selection strategies rely on fixed formulations or parameters shared across all inputs. Hence, they cannot adapt to differences in feature distributions, texture or discriminative granularity across images and domains.These limitations motivate ADAPT-CAM, which is detailed in Section~\ref{methods}.

\section{Methodology}\label{methods}

Motivated by the limitations identified in the previous section, we present ADAPT-CAM, a novel gradient-based, parameter-adaptive CAM framework designed to improve localization precision and stability. The method combines complementary signals to generate sharper and more semantically consistent CAMs. 

\begin{figure*}
   \centering
   \includegraphics[width=0.8\textwidth]{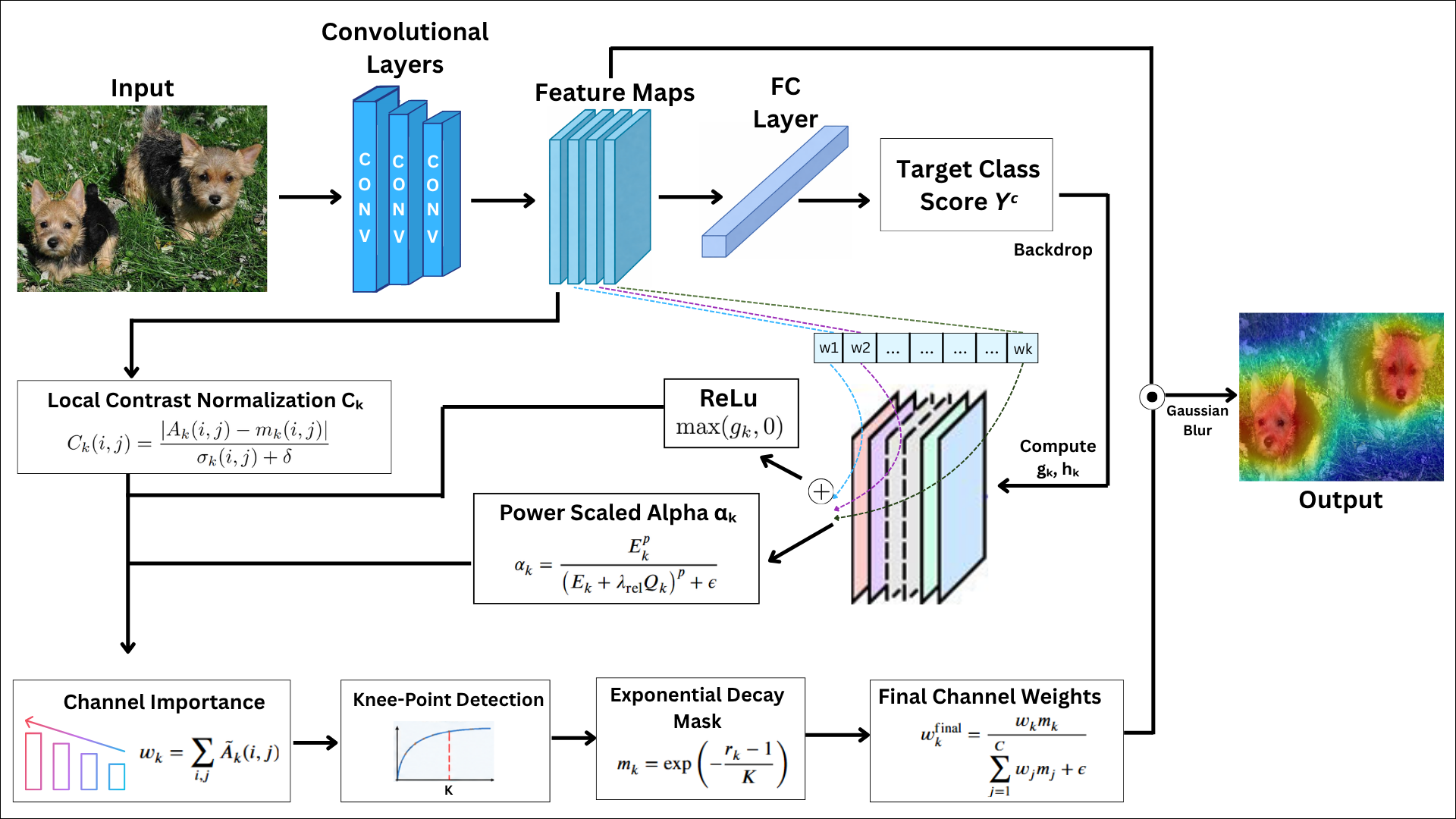}
   \caption{Overview of the proposed parameter-adaptive CAM method}
   \label{fig:proposed_method_pipeline}
\end{figure*}

\subsection{Overall Pipeline of ADAPT-CAM}

Given an input image and a class of interest (e.g., ‘Norwich Terrier’), we forward propagate the image through the trained CNN to obtain a set of non-class conditioned convolutional feature maps. We then compute first-, second-order gradients of the target class score with respect to these feature maps. Based on these gradients, adaptive power-scaled channel weights $\alpha_k$ are computed to quantify the contribution of each channel. In parallel, adaptive LCN is applied to enhance discriminative spatial responses within each feature map with it's neighborhood size determined by the effective spatial distribution of gradient-weighted activations. The resulting LCN maps are combined with the rectified first-order gradients and their corresponding $\alpha_k$ weights to produce spatially resolved channel contributions. An adaptive soft channel selection mechanism is subsequently applied to identify the dominant contribution transition through cumulative contribution-based knee-point detection and progressively attenuate lower-ranked channels using a knee-guided exponential rank-based mask. Finally, the selected activation maps are aggregated, normalized and smoothed using a Gaussian filter to generate the final CAM.

Figure~\ref{fig:proposed_method_pipeline} illustrates the complete workflow of the proposed CAM method. 

\subsection{Power-Scaled Alpha ($\alpha$) Computation for Channel Weighting}

Before defining our channel-wise weighting $\alpha_k$, we briefly explain the rationale for the adopted gradient formulation. Gradient-based CAM methods derive channel importance from class-specific gradient responses, with different formulations emphasizing different aspects of gradient information. For example, Grad-CAM \cite{b12} employs first-order gradients, whereas Grad-CAM++ \cite{b13} incorporates higher-order derivative information into its weighting formulation. In this work, we consider first- and second-order derivatives to capture complementary information about the target class score. This is because the objective is to complement first-order sensitivity with a curvature-related measure, rather than to model successive variations of curvature. Incorporating additional derivative orders would increase computational complexity without being necessary for constructing the proposed weighting formulation. First-order gradients quantify the local sensitivity of the target score to feature activations, whereas second-order derivatives characterize how this sensitivity changes with respect to activation variations. Their combination therefore provides information about both gradient magnitude and activation-dependent curvature.

Let $s^c$ denote post-softmax probability of the target class $c$ and let $A_k(i,j)$ denote the activation of channel $k$ at spatial location $(i,j)$ in the last convolutional feature map. The first-order gradient is denoted by $g_k$. For the second-order term, we use the curvature response $h_k(i,j)=\partial\big(\sum_{k',i',j'} g_{k'}(i',j')\big)/\partial A_k(i,j)$, which is the Hessian of $s^c$ with respect to the feature map applied to a unit vector. This is a spatially and channel-aggregated curvature proxy rather than a full or diagonal Hessian, and it requires only one additional backward pass. This keeps the cost close to that of first-order methods and reflects how the gradient field around each activation changes under the nonlinearities between the feature map and the class score. To characterize the contribution of each channel from complementary gradient perspectives, we compute two channel-wise statistics: gradient energy $E_k$ and activation-weighted curvature contribution $Q_k$:

\begin{equation}
E_k=\sum_{i,j}g_k(i,j)^2,
\qquad
Q_k=\sum_{i,j}|h_k(i,j)||A_k(i,j)|
\end{equation}

The gradient energy $E_k$ quantifies the accumulated first-order sensitivity of the target score to each channel, with squaring preventing positive and negative gradient responses from cancelling one another. However, a first-order gradient is only a local linear estimate of how the score responds to a channel, and its reliability depends on how quickly that response changes around the current activations. To capture this, $Q_k$ combines the magnitude of the second-order response with the corresponding activations, so that the curvature contribution depends on the features present in each channel rather than treating second-order responses independently of their activations.

We use these two statistics to formulate a curvature-modulated channel weighting:

\begin{equation}
\alpha_k =
\frac{E_k^p}
{\left(E_k+\lambda_{\mathrm{rel}}Q_k\right)^p+\epsilon}
\end{equation}

Here, \(E_k\) denotes the gradient energy of channel \(k\), while \(Q_k\) represents its activation-weighted curvature contribution. The resulting coefficient \(\alpha_k\) acts as a curvature-dependent modulation of the first-order channel contribution rather than an independent measure of channel importance. Specifically, it can be expressed approximately as \[\alpha_k \approx \left(1+\lambda_{\mathrm{rel}}\frac{Q_k}{E_k}\right)^{-p},\] making its dependence on the relative curvature contribution explicit. Thus, \(\alpha_k\) approaches one when the curvature contribution is small relative to the gradient energy and decreases as the relative curvature contribution increases. A relatively large \(Q_k/E_k\) indicates a stronger second-order response compared with the channel's first-order gradient energy, suggesting greater variation in the local gradient field around the current activations. Under this formulation, such channels are attenuated because their first-order gradients may provide a less stable local approximation of the target score's response. Conversely, channels with relatively small curvature contributions retain greater influence in the subsequent channel aggregation. Importantly, the overall first-order response magnitude is not determined by \(\alpha_k\) alone; it enters separately through the first-order gradient term in the combined channel contribution, which is further modulated by the local contrast response. This treatment differs from Grad-CAM++ \cite{b13}, where higher-order derivatives refine the spatial weighting of gradients within a channel. In our formulation, curvature instead regulates the relative influence of individual channels by modulating their first-order contributions. The parameter \(\lambda_{\mathrm{rel}}\) controls the relative influence of the curvature contribution, while \(p\) determines the strength of its nonlinear attenuation. Together, these parameters provide an adaptive mechanism for balancing first-order channel responses against their associated higher-order behaviour.

Importantly, neither parameter is a manually selected constant. Both are estimated separately for each input image from its channel-wise statistics, although they capture different characteristics of the channel distribution. The parameter $\lambda_{\mathrm{rel}}$ is obtained as the energy-weighted geometric mean of the channel-wise ratios $E_k/Q_k$. Let $\mathcal{V}$ denote the set of channels with non-negligible gradient energy and non-negligible curvature contribution, and let $w_k=E_k/\sum_{j\in\mathcal{V}}E_j$ represent the normalized share of gradient energy carried by channel $k\in\mathcal{V}$. Then,

\begin{equation}
\lambda_{\mathrm{rel}}=\exp\!\Big(\sum_{k\in\mathcal{V}} w_k \ln\frac{E_k}{Q_k}\Big).
\end{equation}

This provides an image-dependent calibration between gradient energy and curvature by aggregating their relative contributions across valid channels. Channels with larger gradient energy exert greater influence on this estimate, while the geometric mean limits the influence of channels with extreme $E_k/Q_k$ ratios. If no channel satisfies the validity condition, $\lambda_{\mathrm{rel}}$ is set to zero, resulting in no curvature modulation. The exponent $p$, in contrast, is estimated exclusively from the distribution of channel gradient energies. We sort the energies of channels with non-negligible gradient energy in descending order, $E_{(1)}\ge E_{(2)}\ge\dots$, and fit a straight line to $\ln E_{(r)}$ against $\ln r$ using least squares. The decay rate $s$ is defined as the negative of the fitted slope and clipped at zero to ensure $s\geq 0$. The exponent is subsequently determined as

\begin{equation}
p=\frac{s}{1+s},
\end{equation}

which maps the estimated decay rate to the interval $[0,1)$. A steeper decay indicates that gradient information is concentrated in fewer channels, producing a value of $p$ closer to one, whereas a flatter distribution produces a value closer to zero. Together, these image-dependent parameters adapt the relative curvature modulation and its nonlinear attenuation according to the gradient characteristics of each input image. Consequently, the resulting weighting regulates channel contributions by accounting for their first-order sensitivity and relative activation-weighted curvature. The resulting channel-wise coefficients $\alpha_k$ are reshaped and broadcast across the spatial dimensions of the feature maps. They subsequently modulate the spatial first-order gradient responses in conjunction with the corresponding local contrast contributions, forming the combined channel contributions used in the subsequent adaptive soft channel selection stage. Thus, $\alpha_k$ serves as a curvature-modulation factor within the channel-weighting formulation rather than directly determining the final class activation map.

\subsection{Local Contrast Normalization and Combined Channel Weighting}

Although the channel-wise weighting captures the relative importance of feature channels, it does not explicitly account for the spatial distribution of discriminative information within individual feature maps. Moreover, relying exclusively on gradient responses may overlook locally distinctive activation patterns that provide useful structural information. To address these limitations, we introduce adaptive Local Contrast Normalization (LCN) followed by a spatially resolved combination of channel importance, positive gradients, and local contrast.

For adaptive neighborhood estimation, we first construct a gradient-weighted activation map by aggregating the absolute products of activations and their corresponding first-order gradients across channels:

\begin{equation}
M(i,j) = \sum_{k=1}^{K} |g_k(i,j)A_k(i,j)|
\end{equation}

This map represents the spatial distribution of activation responses weighted by their gradient magnitudes, providing an indication of where feature activity is associated with changes in the target prediction. We estimate the effective spatial extent of this distribution using the participation ratio:

\begin{equation}
R_{\mathrm{eff}} =
\frac{\left(\sum_{i,j}M(i,j)\right)^2}
{\sum_{i,j}M(i,j)^2+\epsilon}
\end{equation}

The participation ratio estimates the number of spatial locations effectively contributing to the distribution: concentrated responses produce a smaller value, whereas broadly distributed responses produce a larger one. We convert this effective area into an equivalent spatial radius and determine the LCN window size as:

\begin{equation}
s = \operatorname{Odd}\left(
\operatorname{Clip}\left(
\operatorname{Round}\left(
2\sqrt{\frac{R_{\mathrm{eff}}}{\pi}}
\right),3,s_{\max}
\right)\right)
\end{equation}

where $s_{\max}$ is the largest odd window size not exceeding the smaller feature-map dimension. The window is constrained to a minimum of three pixels, with a $1\times1$ window used when the feature-map dimensions are insufficient. This image-dependent estimation avoids imposing the same normalization scale on inputs with different spatial distributions of gradient-weighted activations.

Using this neighborhood, we calculate the local mean $m_k(i,j)$ and standard deviation $\sigma_k(i,j)$ of each feature map through local averaging. The resulting local contrast is defined as:

\begin{equation}
C_k(i,j) =
\frac{|A_k(i,j)-m_k(i,j)|}
{\sigma_k(i,j)+10^{-4}}
\end{equation}

where the local standard deviation is obtained from the local second moment, with its variance bounded below by $10^{-12}$ for numerical stability. By measuring activation deviations relative to their surrounding local variation, this normalization emphasizes spatially distinctive responses while reducing the relative contribution of activations that are homogeneous within their neighborhoods. Consequently, the adaptive neighborhood allows local contrast to be evaluated at a spatial scale determined by the effective distribution of gradient-weighted activations.

We subsequently combine the previously computed channel-wise weighting coefficient \(\alpha_k\) with the positive first-order gradients and local contrast map to obtain the combined per-channel spatial contribution:

\begin{equation}
\tilde{A}_k(i,j) =
\alpha_k \cdot
\operatorname{ReLU}(g_k(i,j))
\cdot C_k(i,j)
\label{eq:att_weight}
\end{equation}

where \(\alpha_k\) represents the channel-level importance derived from the power-scaled combination of gradient energy and curvature information, \(g_k(i,j)\) denotes the spatially varying first-order gradient, and \(C_k(i,j)\) captures the local structural contrast within the corresponding feature map.

The three components in Eq.~(\ref{eq:att_weight}) provide complementary information. First, \(\alpha_k\) determines the relative contribution of each channel by incorporating its gradient energy and curvature-weighted activation contribution. Second, \(\operatorname{ReLU}(g_k(i,j))\) preserves positive gradient responses at individual spatial locations, emphasizing regions where increasing the corresponding feature activation supports the target-class score. Third, \(C_k(i,j)\) enhances local deviations from the surrounding activation distribution, allowing subtle but discriminative structural variations to contribute to the resulting representation.

Consequently, the combined weighting incorporates channel-level importance, spatially selective class relevance, and local structural contrast within a unified formulation. This provides a more comprehensive representation of discriminative feature responses than relying on channel-level gradient weighting or spatial gradients alone.

\subsection{Adaptive Soft Channel Selection}

Following the integration of channel-level importance, spatial gradient sensitivity, and local contrast, the resulting spatial contributions provide a refined representation of class-relevant information within each feature channel. However, retaining all channels without considering their relative contributions may allow weaker channels to introduce less relevant information into the final CAM. We therefore introduce an adaptive soft channel selection mechanism to prioritize dominant channels while progressively suppressing lower-ranked contributions. Unlike conventional Top-$K$ selection, which requires a predefined channel count and completely discards the remaining channels, our approach determines the dominant channel group from the contribution distribution of each input and continuously attenuates less-contributing channels.

Given the combined spatial contributions $\tilde{A}_k(i,j)$ from Eq.~(\ref{eq:att_weight}), we calculate the importance of each channel through spatial aggregation:

\begin{equation}
w_k = \sum_{i,j}\tilde{A}_k(i,j)
\end{equation}

These scores represent the overall contribution of each channel after incorporating gradient-based importance, spatial class sensitivity, and local contrast. Only channels with positive importance ($w_k>0$) are considered, excluding channels that provide no positive class-relevant contribution. The remaining channels are sorted in descending order of importance, with $r_k$ denoting the rank of channel $k$. Their normalized cumulative contributions are then examined to identify the selection boundary.

Specifically, the adaptive selection boundary $K$ is determined through knee-point detection as the rank corresponding to the maximum perpendicular distance between the cumulative contribution curve and the straight line connecting its endpoints. This identifies the transition between dominant and relatively less-contributing channels without requiring a predefined selection count. If the cumulative curve is effectively linear and no distinguishable knee exists, $K$ is set to the total number of active channels.

To progressively suppress lower-ranked contributions without imposing a hard selection cutoff, we employ a continuous exponential rank-based selection mechanism:

\begin{equation}
m_k = \exp\left(-\frac{r_k-1}{K}\right).
\end{equation}

The highest-ranked channel receives a selection coefficient of one, while the coefficients of subsequent channels decay exponentially according to their ranks. The decay scale is determined entirely by the image-adaptive knee $K$. Consequently, contribution distributions with a smaller detected knee produce faster attenuation of lower-ranked channels, whereas distributions with a larger knee result in a more gradual decay. This allows the selection mechanism to adapt to the contribution distribution of each input while avoiding abrupt channel removal, predefined channel counts, or manually specified attenuation parameters.

Finally, the selection coefficients are multiplied by their corresponding original channel importance scores to obtain the normalized final channel weights:

\begin{equation}
w_k^{\mathrm{final}} =
\frac{w_k m_k}
{\displaystyle\sum_{j=1}^{C}w_jm_j+\epsilon},
\end{equation}

where $C$ denotes the total number of feature channels and $\epsilon$ is a small numerical constant for numerical stability. Channels with zero importance receive zero weight, whereas all positive-importance channels retain nonzero contributions through the continuous selection mechanism. The resulting normalized weights preserve the relative importance of dominant channels while progressively reducing the influence of weaker contributions. These weights are subsequently used to aggregate the feature maps and generate the final class activation map.

\subsection{Final Class Activation Map Generation}
The resulting weights are normalized and used to combine the original feature maps, producing the final CAM:

\begin{equation}
L_{\mathrm{CAM}}(i,j) =
\sum_{k=1}^{C}
\frac{w_k^{\mathrm{final}}}
{\sum_{\ell=1}^{C}w_\ell^{\mathrm{final}}+\epsilon}
A_k(i,j).
\end{equation}

The resulting activation map is subsequently smoothed and resized to the input resolution. Gaussian smoothing \cite{b37} is applied as a spatial post-processing operation because the combined effect of LCN and channel selection produces highly localized and specific activation responses.
It attenuates abrupt local variations by incorporating neighbouring activation values according to their spatial proximity, thereby promoting spatial continuity and producing a more coherent representation. ADAPT-CAM employs mild smoothing of $\sigma = 0.5$ to produce a smoother representation while retaining the overall activation structure. This operation does not alter the channel-selection weights or their relative importance and only refines the spatial distribution of the final activation map. The smoothed map is then finally resized to the input resolution.

\subsection{Experimental Setup}
\label{expsetup}

ADAPT-CAM is evaluated across multiple CNN and Transformer architectures on two image classification datasets: ImageNet \cite{b31}  and Oxford 102 Flowers \cite{b32}. For ImageNet \cite{b31}, experiments are conducted using DenseNet121 \cite{b33}, ResNet-50 \cite{b34}, DINOv2 \cite{b41}, and ViT-B/16 \cite{b42} on 500 randomly selected validation images. For Oxford 102 Flowers \cite{b32}, experiments are conducted using ResNet-50 \cite{b34} and ViT-B/16 \cite{b42} on 500 randomly selected images to further assess performance on a fine-grained classification dataset. The final convolutional layer is used as the target layer in CNN architectures for all CAM methods. In transformer architectures, the features are taken from the final encoder block output, CLS token excluded and the remaining patch tokens reshaped into their respective spatial grids. The 768-dimensional token embeddings are treated as feature channels with each embedding dimension forming a spatial feature map across the patch grid. This evaluation covers both conventional CNN architectures as well as Vision Transformers. Although ADAPT-CAM was initially designed and formulated with CNNs in mind, its evaluation on Transformer architectures also investigates its applicability beyond CNN-based models. All baseline methods are evaluated using their officially recommended parameter settings for fair comparison. The explanation quality is quantitatively evaluated using four established metrics: Insertion AUC, Deletion AUC \cite{b39}, Average Drop, and Average Increase \cite{b13}. The study follows standard evaluation protocol for these metrics using 50 steps and the softmax probability of the model-predicted class. Higher Insertion AUC and Average Increase, together with lower Deletion AUC and Average Drop, indicate more faithful explanations. All metrics are reported as mean $\pm$ standard error of the mean (SEM), calculated over the 500 evaluated images. Lower SEM values indicate more precise mean estimates whereas higher values indicate some level of uncertainty in the estimated means.

\section{Results and Discussion}\label{results}
This section reports the quantitative and qualitative evaluation of ADAPT-CAM against established CAM baselines, following the protocol described in Section~\ref{expsetup}. 

\subsection{Quantitative Results on ImageNet}
The ImageNet \cite{b31} experiments evaluate ADAPT-CAM on DenseNet121 \cite{b33}, ResNet-50 \cite{b34}, DINOv2 \cite{b41}, and ViT-B/16 \cite{b42}. This provides an assessment across conventional convolutional architectures and Vision Transformers, including both supervised and self-supervised pretrained models. 

Tables~\ref{tbl:imagenet_cnn} reports the results for the two convolutional backbones. 

\begin{table*}[t]
\centering
\caption{Evaluation of ADAPT-CAM and baseline methods on ImageNet using DenseNet-121 and ResNet-50 (mean $\pm$ SEM over 500 images)}
\label{tbl:imagenet_cnn}
\begin{tabular}{llcccc}
\toprule
Architecture & Method & Insertion $\uparrow$ & Deletion $\downarrow$ & Avg Drop (\%) $\downarrow$ & Avg Increase (\%) $\uparrow$ \\
\midrule

\multirow{7}{*}{DenseNet}
& Grad-CAM \cite{b12} & 0.517 $\pm$ 0.0120 & 0.097 $\pm$ 0.0044 & 52.785 $\pm$ 1.8957 & 10.80 $\pm$ 1.39 \\
& Grad-CAM++ \cite{b13} & 0.578 $\pm$ 0.0121 & 0.094 $\pm$ 0.0042 & 38.860 $\pm$ 1.8024 & 18.60 $\pm$ 1.74 \\
& Smooth Grad-CAM++ \cite{b14} & 0.510 $\pm$ 0.0121 & 0.127 $\pm$ 0.0061 & 47.510 $\pm$ 1.6147 & 6.60 $\pm$ 1.11 \\
& XGrad-CAM \cite{b18} & 0.444 $\pm$ 0.0131 & 0.184 $\pm$ 0.0082 & 59.005 $\pm$ 1.5634 & 3.60 $\pm$ 0.83 \\
& Score-CAM \cite{b15} & 0.471 $\pm$ 0.0102 & 0.233 $\pm$ 0.0103 & 41.795 $\pm$ 0.9252 & 7.80 $\pm$ 1.20 \\
& EigenCAM \cite{b16} & 0.526 $\pm$ 0.0136 & 0.092 $\pm$ 0.0051 & 48.967 $\pm$ 1.4472 & 3.10 $\pm$ 0.76 \\
& \textbf{ADAPT-CAM (Ours)} & \textbf{0.582 $\pm$ 0.0119} & \textbf{0.089 $\pm$ 0.0046} & \textbf{16.756 $\pm$ 1.1258} & \textbf{34.60 $\pm$ 2.13} \\

\midrule

\multirow{7}{*}{ResNet}
& Grad-CAM \cite{b12} & 0.508 $\pm$ 0.013 & 0.093 $\pm$ 0.005 & 54.105 $\pm$ 1.439 & 7.85 $\pm$ 1.20 \\
& Grad-CAM++ \cite{b13} & 0.506 $\pm$ 0.013 & 0.122 $\pm$ 0.004 & 41.195 $\pm$ 1.706 & 10.40 $\pm$ 1.37 \\
& Smooth Grad-CAM++ \cite{b14} & 0.430 $\pm$ 0.012 & 0.136 $\pm$ 0.006 & 53.023 $\pm$ 0.975 & 11.89 $\pm$ 1.45 \\
& XGrad-CAM \cite{b18} & 0.498 $\pm$ 0.013 & 0.122 $\pm$ 0.006 & 64.939 $\pm$ 1.447 & 5.90 $\pm$ 1.05 \\
& Score-CAM \cite{b15} & 0.523 $\pm$ 0.013 & 0.116 $\pm$ 0.005 & 48.928 $\pm$ 1.302 & 12.75 $\pm$ 1.49 \\
& EigenCAM \cite{b16} & 0.458 $\pm$ 0.013 & 0.088 $\pm$ 0.005 & 55.965 $\pm$ 1.115 & 5.80 $\pm$ 1.05 \\
& \textbf{ADAPT-CAM (Ours)} & \textbf{0.530 $\pm$ 0.012} & \textbf{0.084 $\pm$ 0.004} & \textbf{19.120 $\pm$ 1.112} & \textbf{25.80 $\pm$ 1.96} \\
\bottomrule
\end{tabular}
\end{table*}

In terms of Insertion AUC and Deletion AUC, ADAPT-CAM demonstrates competitive performance with existing methods across both architectures. Its results remain close to those of the strongest competing approaches, including Grad-CAM++ \cite{b13} and EigenCAM \cite{b16} for DenseNet-121 \cite{b33}, and Score-CAM \cite{b15} and EigenCAM \cite{b16} for ResNet-50 \cite{b34}.

This indicates comparable performance under insertion- and deletion-based evaluation. In contrast, a more pronounced distinction is observed in Average Drop and Average Increase. ADAPT-CAM achieves substantially lower Average Drop values of 16.756\% and 19.120\% on DenseNet-121 \cite{b33} and ResNet-50 \cite{b34}, respectively. These are considerably lower than the corresponding results of Grad-CAM++ \cite{b13} which is the closest baseline on both architectures recording 38.860\% and 41.195\%. Similarly, ADAPT-CAM obtains Average Increase values of 34.60\% and 25.80\% as compared with 18.60\% for Grad-CAM++ \cite{b13} on DenseNet-121 \cite{b33} and 12.75\% for Score-CAM \cite{b15} on ResNet-50 \cite{b34}. This consistent pattern across both architectures highlights a particularly substantial performance difference in these two metrics.

The SEM values provide an indication of the precision of the estimated mean performance. For Insertion AUC, ADAPT-CAM exhibits SEM values comparable to those of the competing methods on both architectures indicating similar and more precision in the estimated results. A similar pattern is observed for Deletion AUC. For Average Drop, ADAPT-CAM achieves relatively low SEM values on both backbones. For Average Increase, although its SEM values are comparatively higher than those of the baselines, its mean performance remains considerably higher across both architectures.

Tables~\ref{tbl:imagenet_transformers} reports the results for the two transformer backbones. The comparison set additionally includes methods used for Vision Transformers as well.

\begin{table*}[t]
\centering
\caption{Evaluation of ADAPT-CAM and baseline methods on ImageNet using DINOv2 and ViT-B/16 (mean $\pm$ SEM over 500 images)}
\label{tbl:imagenet_transformers}
\begin{tabular}{llcccc}
\toprule
Architecture & Method & Insertion $\uparrow$ & Deletion $\downarrow$ & Avg Drop (\%) $\downarrow$ & Avg Increase (\%) $\uparrow$ \\
\midrule

\multirow{7}{*}{DINOv2}
& Grad-CAM \cite{b12} & 0.786 $\pm$ 0.012 & 0.533 $\pm$ 0.015 & 13.220 $\pm$ 0.841 & 29.60 $\pm$ 2.04 \\
& Grad-CAM++ \cite{b13} & 0.798 $\pm$ 0.013 & 0.520 $\pm$ 0.015 & 13.390 $\pm$ 0.868 & 25.55 $\pm$ 1.95 \\
& Score-CAM \cite{b15} & 0.736 $\pm$ 0.014 & 0.357 $\pm$ 0.016 & 20.239 $\pm$ 1.233 & 22.87 $\pm$ 1.88 \\
& RISE \cite{b39} & 0.758 $\pm$ 0.011 & 0.315 $\pm$ 0.015 & 17.671 $\pm$ 0.819 & 25.72 $\pm$ 1.95 \\

& Attention Rollout \cite{b43} & 0.840 $\pm$ 0.012 & 0.319 $\pm$ 0.015 & 8.771 $\pm$ 0.728 & 27.90 $\pm$ 2.01 \\
& ViT-CX \cite{b44} & 0.775 $\pm$ 0.014 & 0.451 $\pm$ 0.016 & 16.347 $\pm$ 1.191 & 26.07 $\pm$ 1.96 \\
& \textbf{ADAPT-CAM (Ours)} & \textbf{0.875 $\pm$ 0.016} & \textbf{0.104 $\pm$ 0.006} & \textbf{7.288 $\pm$ 1.104} & \textbf{31.20 $\pm$ 2.07} \\

\midrule

\multirow{7}{*}{ViT-B/16}
& Grad-CAM \cite{b12} & 0.628 $\pm$ 0.018 & 0.585 $\pm$ 0.017 & 30.736 $\pm$ 1.586 & 25.90 $\pm$ 1.96 \\
& Grad-CAM++ \cite{b13} & 0.525 $\pm$ 0.019 & 0.541 $\pm$ 0.019 & 41.848 $\pm$ 1.775 & 18.70 $\pm$ 1.74 \\
& Score-CAM \cite{b15} & 0.642 $\pm$ 0.018 & 0.580 $\pm$ 0.017 & 29.104 $\pm$ 1.580 & 28.95 $\pm$ 2.03 \\
& RISE \cite{b39} & 0.734 $\pm$ 0.011 & 0.484 $\pm$ 0.017 & 20.171 $\pm$ 0.785 & 29.90 $\pm$ 2.05 \\

& Attention Rollout \cite{b43} & 0.727 $\pm$ 0.016 & 0.494 $\pm$ 0.015 & 21.659 $\pm$ 1.165 & 32.10 $\pm$ 2.09 \\
& ViT-CX \cite{b44} & 0.619 $\pm$ 0.016 & 0.544 $\pm$ 0.015 & 31.838 $\pm$ 1.412 & 26.25 $\pm$ 1.97 \\
& \textbf{ADAPT-CAM (Ours)} & \textbf{0.754 $\pm$ 0.017} & \textbf{0.352 $\pm$ 0.016} & \textbf{19.171 $\pm$ 1.687} & \textbf{38.65 $\pm$ 2.18} \\
\bottomrule
\end{tabular}
\end{table*}

In terms of Insertion AUC and Deletion AUC, ADAPT-CAM demonstrates strong performance across both transformer architectures. For DINOv2 \cite{b41}, ADAPT-CAM achieves the highest Insertion AUC of 0.875. This exceeds the strongest baseline of Attention Rollout \cite{b43} which records 0.840. It also obtains a considerably lower Deletion AUC of 0.104 compared with the strongest baseline result of 0.315 from RISE \cite{b39}. A similar pattern is observed with ViT-B/16 \cite{b42}, where ADAPT-CAM achieves the highest Insertion AUC of 0.754, followed by RISE \cite{b39} at 0.734. For Deletion AUC, ADAPT-CAM records 0.352 as compared with 0.484 for RISE \cite{b39} which represents the strongest competing result. These results indicate that ADAPT-CAM maintains competitive performance across both insertion- and deletion-based evaluations with considerable differences in Deletion AUC.

A similar distinction is observed in Average Drop and Average Increase. For DINOv2 \cite{b41}, ADAPT-CAM achieves the lowest Average Drop of 7.288\% which is followed by Attention Rollout \cite{b43} at 8.771\%. It also obtains the highest Average Increase of 31.20\% compared with 29.60\% for Grad-CAM \cite{b12} which is the closest baseline. For ViT-B/16 \cite{b42}, ADAPT-CAM again records the lowest Average Drop of 19.171\% which marginally outperforms RISE \cite{b39} at 20.171\%. Its Average Increase reaches 38.65\% which exceeds Attention Rollout \cite{b43}, the closest baseline at 32.10\%. This consistent pattern across both transformer architectures demonstrates the ability of ADAPT-CAM to achieve lower Average Drop while maintaining higher Average Increase relative to the competing methods.

The SEM values provide an indication of the precision of the estimated mean performance. For Insertion AUC and Average Drop, ADAPT-CAM exhibits competitive SEM values for DINOv2 \cite{b41} and ViT-B/16 \cite{b42}. For Deletion AUC, it achieves comparatively low SEM values of 0.006 on DINOv2 \cite{b41} and 0.016 on ViT-B/16 \cite{b42} indicating relatively precise estimates particularly for DINOv2. For Average Increase, ADAPT-CAM records SEM values of 2.07 and 2.18, which are broadly comparable to those of the competing methods. Its higher mean performance across both backbones is therefore accompanied by a similar level of estimation precision.
The relative performance of the baseline methods varies across architectures. For CNN architectures, Grad-CAM++ \cite{b13} performs competitively on DenseNet-121 \cite{b33}, whereas Score-CAM \cite{b15} achieves comparatively stronger results on ResNet-50 \cite{b34}. EigenCAM \cite{b16} also demonstrates competitive Deletion AUC on both backbones. For transformers, Attention Rollout \cite{b43} achieves the strongest baseline Insertion AUC and Average Drop on DINOv2 \cite{b41}, whereas RISE \cite{b39} demonstrates comparatively stronger Deletion AUC. In contrast, RISE \cite{b39} achieves the strongest baseline Insertion AUC, Deletion AUC and Average Drop on ViT-B/16 \cite{b42}, while Attention Rollout \cite{b43} obtains the highest baseline Average Increase. Such differences may arise from variations in feature representation, gradient propagation, token representations and attention distributions across the architectures, which influence how individual attribution methods capture class-discriminative information.

Moreover, ImageNet \cite{b31} contains a thousand classes with large variation in object scale, texture and background clutter, so a single fixed weighting is unlikely to suit every image. This is reflected in the baselines, whose relative ranking changes with the backbone. On the other hand, ADAPT-CAM estimates its weighting parameters and its LCN neighborhood size for each image, so it does not rely on a setting tuned to a particular backbone. This is consistent with it obtaining the best mean value on every metric across all four backbones.

\subsection{Quantitative Results on Oxford 102 Flowers}

To further investigate the effectiveness of ADAPT-CAM on fine-grained visual recognition, additional experiments are conducted on Oxford 102 Flowers \cite{b32} using ResNet-50 \cite{b34} and ViT-B/16 \cite{b42}. Unlike ImageNet \cite{b31}, which contains a broad range of object categories, Oxford 102 Flowers \cite{b32} focuses on visually similar flower categories requiring explanations to capture subtle discriminative characteristics. Tables~\ref{tbl:flowers_combined} presents the results on the Oxford 102 Flowers \cite{b32} dataset. 

\begin{table*}[t]
\centering
\caption{Evaluation of ADAPT-CAM and baseline methods on Oxford 102 Flowers using ResNet-50 and ViT-B/16 (mean $\pm$ SEM over 500 images)}
\label{tbl:flowers_combined}
\begin{tabular}{llcccc}
\toprule
Architecture & Method & Insertion $\uparrow$ & Deletion $\downarrow$ & Avg Drop (\%) $\downarrow$ & Avg Increase (\%) $\uparrow$ \\
\midrule

\multirow{7}{*}{ResNet-50}
& Grad-CAM \cite{b12} & 0.5127 $\pm$ 0.0151 & 0.2594 $\pm$ 0.0054 & 32.97 $\pm$ 1.687 & 11.12 $\pm$ 1.41 \\
& Grad-CAM++ \cite{b13} & 0.5391 $\pm$ 0.0130 & 0.2479 $\pm$ 0.0068 & 46.44 $\pm$ 1.898 & 14.80 $\pm$ 1.59 \\
& Smooth Grad-CAM++ \cite{b14} & 0.5388 $\pm$ 0.0117 & 0.2647 $\pm$ 0.0067 & 47.46 $\pm$ 1.564 & 5.34 $\pm$ 1.01 \\
& XGrad-CAM \cite{b18} & 0.4923 $\pm$ 0.0133 & 0.3181 $\pm$ 0.0059 & 62.62 $\pm$ 1.762 & 4.22 $\pm$ 0.90 \\
& Score-CAM \cite{b15} & 0.5453 $\pm$ 0.0115 & 0.2260 $\pm$ 0.0059 & 38.44 $\pm$ 1.768 & 8.70 $\pm$ 1.26 \\
& EigenCAM \cite{b16} & 0.5338 $\pm$ 0.0089 & 0.2880 $\pm$ 0.0057 & 31.33 $\pm$ 1.512 & 3.80 $\pm$ 0.86 \\
& \textbf{ADAPT-CAM (Ours)} & \textbf{0.5567 $\pm$ 0.0098} & \textbf{0.2122 $\pm$ 0.0038} & \textbf{21.84 $\pm$ 1.447} & \textbf{16.12 $\pm$ 1.64} \\
\midrule

\multirow{7}{*}{ViT-B/16}
& Grad-CAM \cite{b12} & 0.5487 $\pm$ 0.0110 & 0.468 $\pm$ 0.016 & 31.24 $\pm$ 1.486 & 9.89 $\pm$ 1.34 \\
& Grad-CAM++ \cite{b13} & 0.5597 $\pm$ 0.0120 & 0.487 $\pm$ 0.016 & 29.18 $\pm$ 1.676 & 10.42 $\pm$ 1.37 \\
& Score-CAM \cite{b15} & 0.6527 $\pm$ 0.0124 & 0.551 $\pm$ 0.016 & 19.89 $\pm$ 1.470 & 17.87 $\pm$ 1.71 \\
& RISE \cite{b39} & 0.7743 $\pm$ 0.0079 & \textbf{0.364 $\pm$ 0.015} & 17.68 $\pm$ 0.985 & 15.16 $\pm$ 1.60 \\
& Attention Rollout \cite{b43} & 0.8037 $\pm$ 0.0103 & 0.409 $\pm$ 0.017 & 15.35 $\pm$ 1.176 & 22.72 $\pm$ 1.87 \\
& ViT-CX \cite{b44} & 0.7577 $\pm$ 0.0100 & 0.441 $\pm$ 0.017 & 18.71 $\pm$ 1.532 & 21.89 $\pm$ 1.85 \\
& \textbf{ADAPT-CAM (Ours)} & \textbf{0.8248 $\pm$ 0.0056} & 0.393 $\pm$ 0.012 & \textbf{11.171 $\pm$ 1.127} & \textbf{27.89 $\pm$ 2.01} \\

\bottomrule
\end{tabular}
\end{table*}

For ResNet-50 \cite{b34}, ADAPT-CAM demonstrates the stronger overall performance across all four evaluation metrics. It achieves competitive Insertion AUC (0.5567) and Deletion AUC (0.2122). A more pronounced distinction is observed in Average Drop and Average Increase. ADAPT-CAM achieves the lowest Average Drop of 21.84\% which is considerably lower than Eigen-CAM \cite{b16} at 31.33\% representing the strongest baseline result. Similarly, it obtains the highest Average Increase of 16.12\% which exceeds the closest competing method (Grad-CAM++ \cite{b13}) at 14.80\%. 

For ViT-B/16 \cite{b42}, ADAPT-CAM achieves the strongest results in Insertion AUC, Average Drop and Average Increase, while RISE \cite{b39} demonstrates the strongest Deletion AUC. In terms of Insertion AUC, ADAPT-CAM records the highest value of 0.8248, followed by Attention Rollout \cite{b43} at 0.8037. For Deletion AUC, however, RISE \cite{b39} achieves the lowest value of 0.364 compared with 0.393 for ADAPT-CAM. Although ADAPT-CAM does not achieve the strongest result for this metric, its performance remains competitive with the leading baseline method. A similar distinction is observed in Average Drop and Average Increase. ADAPT-CAM achieves the lowest Average Drop of 11.171\%, outperforming Attention Rollout \cite{b43} at 15.35\% which is the closest baseline. It also obtains the highest Average Increase of 27.89\%, exceeding Attention Rollout \cite{b43} at 22.72\%. These results demonstrate that ADAPT-CAM provides particularly strong performance across the three metrics, while maintaining competitive Deletion AUC on ViT-B/16 \cite{b42}.

The SEM values provide further insight on the precision of the performance of methods across both architectures. For ResNet-50 \cite{b34}, ADAPT-CAM exhibits a relatively low SEM of 0.0098 for Insertion AUC, comparable to EigenCAM \cite{b16} at 0.0089. It achieves the lowest SEM for Deletion AUC (0.0038) and Average Drop (1.447), while its Average Increase SEM (1.64) is comparable to Grad-CAM++ \cite{b13} (1.59). For ViT-B/16 \cite{b42}, ADAPT-CAM records the lowest SEM for Insertion AUC (0.0056) and Deletion AUC (0.012). Its Average Drop SEM (1.127) is the second-lowest after RISE \cite{b39} (0.985), whereas its Average Increase SEM (2.01) is higher than those of the competing methods. Overall, these values indicate relatively precise mean estimates, although the degree of precision varies across metrics.

On ResNet-50 \cite{b34}, Score-CAM \cite{b15} is the strongest baseline in Insertion and Deletion AUC, EigenCAM \cite{b16} in Average Drop, and Grad-CAM++ \cite{b13} in Average Increase. On ViT-B/16 \cite{b42}, the strongest baselines are instead transformer-oriented methods: Attention Rollout \cite{b43} in Insertion AUC, Average Drop and Average Increase, and RISE \cite{b39} in Deletion AUC, with ViT-CX \cite{b44} also competitive in Average Increase. These rankings do not carry over between backbones. Score-CAM \cite{b15} records the weakest Deletion AUC on ViT-B/16 (0.551), and Grad-CAM++ \cite{b13} reaches an Average Increase of only 10.42\% compared with 14.80\% on ResNet-50 \cite{b34}.

Oxford 102 Flowers consists of visually similar flower categories in which class-discriminative evidence may lie in subtle details such as petal shape, colour and texture. This provides a useful context for interpreting the results of EigenCAM \cite{b16} on ResNet-50 \cite{b34}. As a class-agnostic method, it highlights the dominant flower regardless of the target class, which retains enough evidence to give the strongest baseline Average Drop (31.33\%), but it has the lowest Average Increase (3.80\%) because it does not sharpen class-specific evidence. In contrast, ADAPT-CAM combines local contrast with spatial gradient information and adaptive soft channel selection which allows its explanations to account for both local activation variations and channel-level importance. Its image-adaptive weighting and neighborhood estimation further allow these components to respond to the characteristics of individual inputs. This is consistent with it obtaining the best Insertion AUC, the lowest Average Drop and the highest Average Increase on both architectures, and the best Deletion AUC on ResNet-50 \cite{b34}. The only exception is Deletion AUC on ViT-B/16 \cite{b42}, where RISE \cite{b39} is stronger (0.364 against 0.393). Overall, the results demonstrate strong performance across both CNN and Transformer architectures.

\subsection{Qualitative Evaluation}
\label{qualitative_results}

To complement the quantitative evaluation, qualitative comparisons of the generated class activation maps are presented for the different architectures and datasets. These visualizations provide an assessment of the spatial localization characteristics of ADAPT-CAM in comparison with the baseline methods. 

\begin{figure*}
    \centering
    \includegraphics[width=0.8\textwidth]{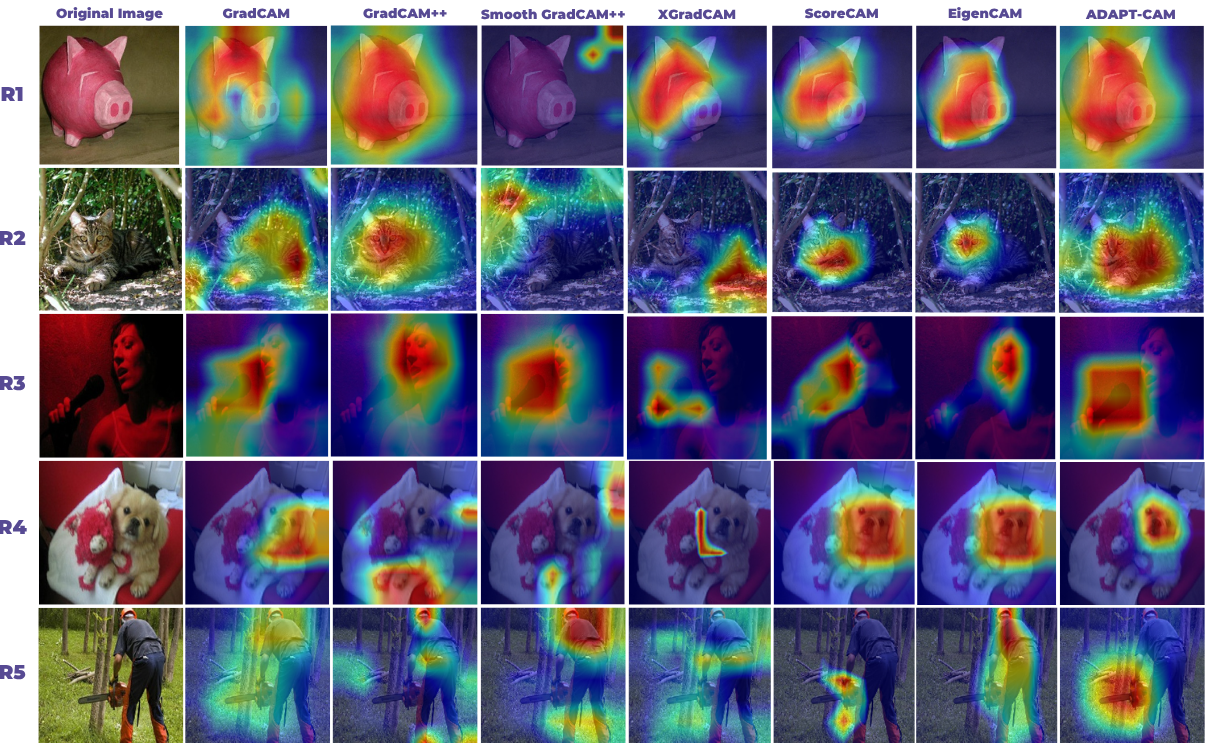}
    \caption{Qualitative comparison on ImageNet~\cite{b31} for DenseNet121~\cite{b33}}
    \label{figure2}
\end{figure*}

For ImageNet \cite{b31}, the qualitative comparisons are first presented for DenseNet121 \cite{b33} in Fig.~\ref{figure2}. Here, Grad-CAM~\cite{b12}, Grad-CAM++~\cite{b13} and Score-CAM~\cite{b15} generally cover the principal object with varying spread. Whereas, XGrad-CAM~\cite{b18} is less stable (e.g., R2, and a narrow artifact in R4). Smooth Grad-CAM++~\cite{b14} frequently misses the object (e.g. R1, R2) and EigenCAM~\cite{b16} behaves much better yet still misses objects sometimes as shown in R5. ADAPT-CAM consistently covers the pig (R1), cat (R2), and puppy (R4), with its tightest localization in R4. In R1 and R2 its extent is comparable to the strongest baselines.

\begin{figure*}
    \centering
    \includegraphics[width=0.8\textwidth]{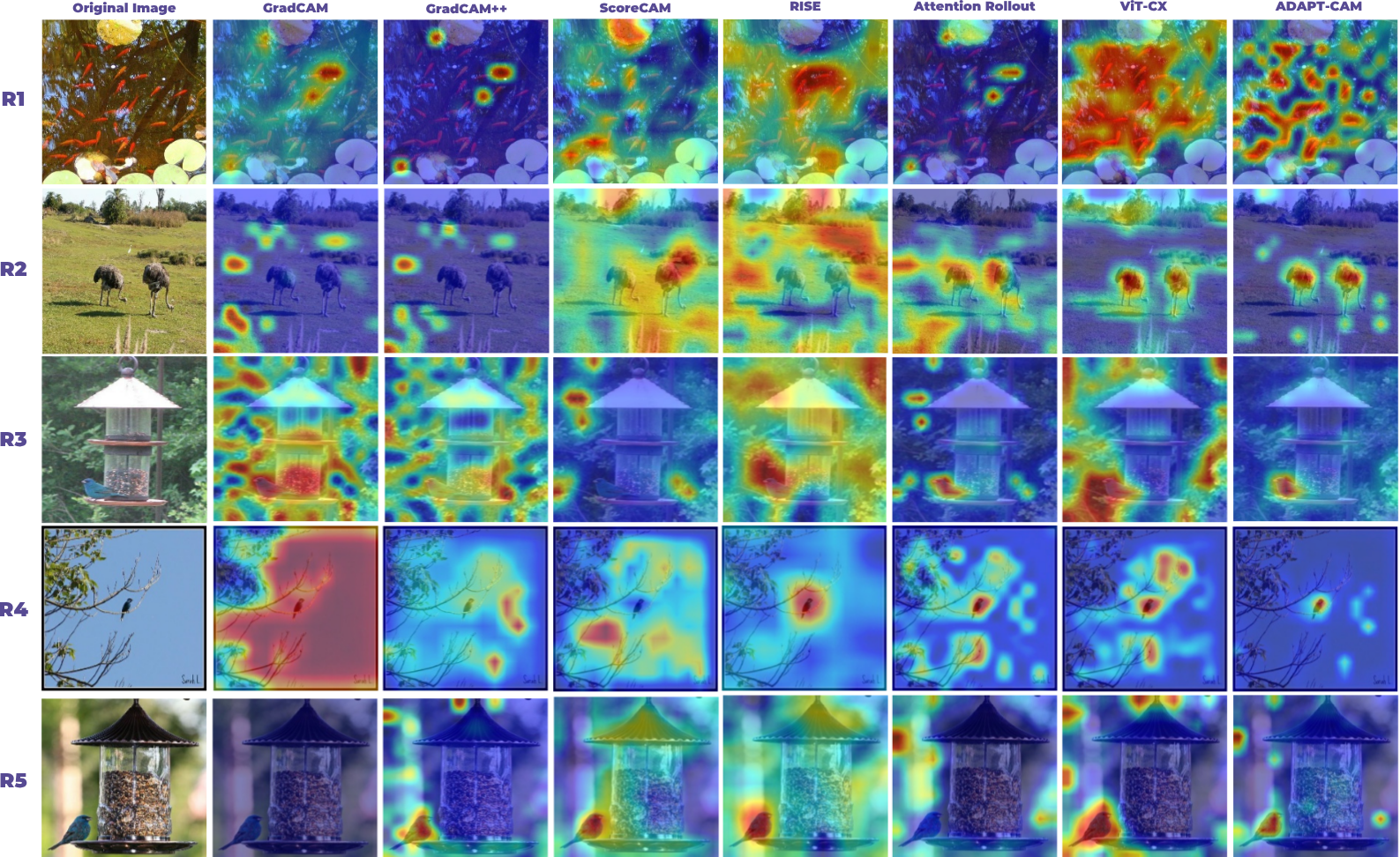}
    \caption{Qualitative comparison on ImageNet~\cite{b31} for ViT-B/16~\cite{b42}}
    \label{figure3}
\end{figure*}

The comparisons are then extended to ViT-B/16~\cite{b42} as shown in Fig.~\ref{figure3}.Grad-CAM~\cite{b12} produces coarse, broad activations (e.g., R4), while Grad-CAM++~\cite{b13} is more compact but often lands on background rather than the object (R2). Score-CAM~\cite{b15} and RISE~\cite{b39} spread activation over large background areas. Attention Rollout~\cite{b43} and ViT-CX~\cite{b44} highlight the ostriches (R2) and the bird (R4) but with more diffuse or fragmented responses elsewhere. ADAPT-CAM concentrates activation on the target objects in R2--R5 with comparatively little background response; in the cluttered underwater scene (R1), its response is fragmented across several regions.

\begin{figure*}
    \centering
    \includegraphics[width=0.8\textwidth]{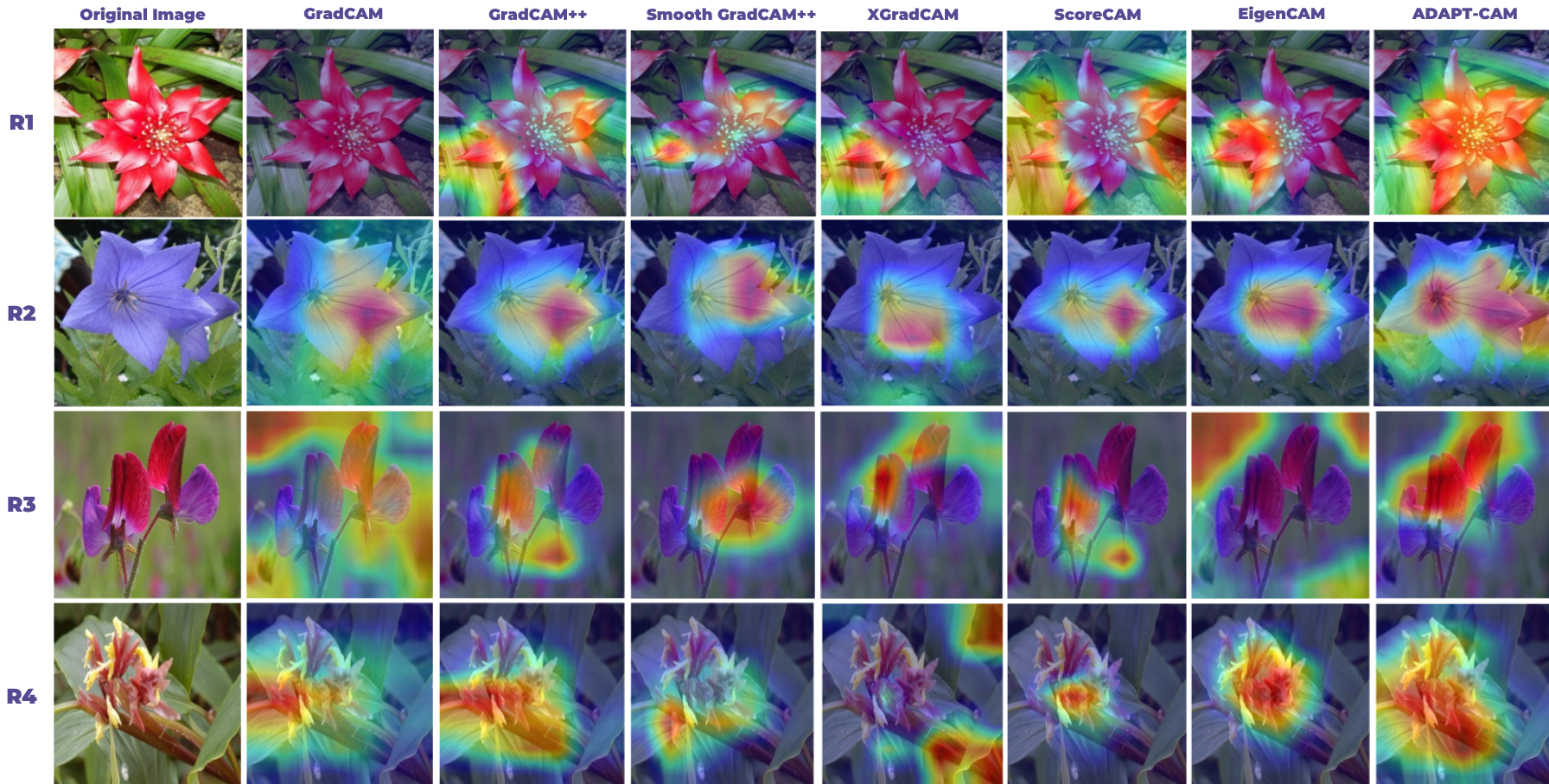}
    \caption{Qualitative comparison on Oxford 102 Flowers~\cite{b32} for ResNet-50~\cite{b34}}
    \label{figure4}
\end{figure*}

\begin{figure*}
    \centering
    \includegraphics[width=0.8\textwidth]{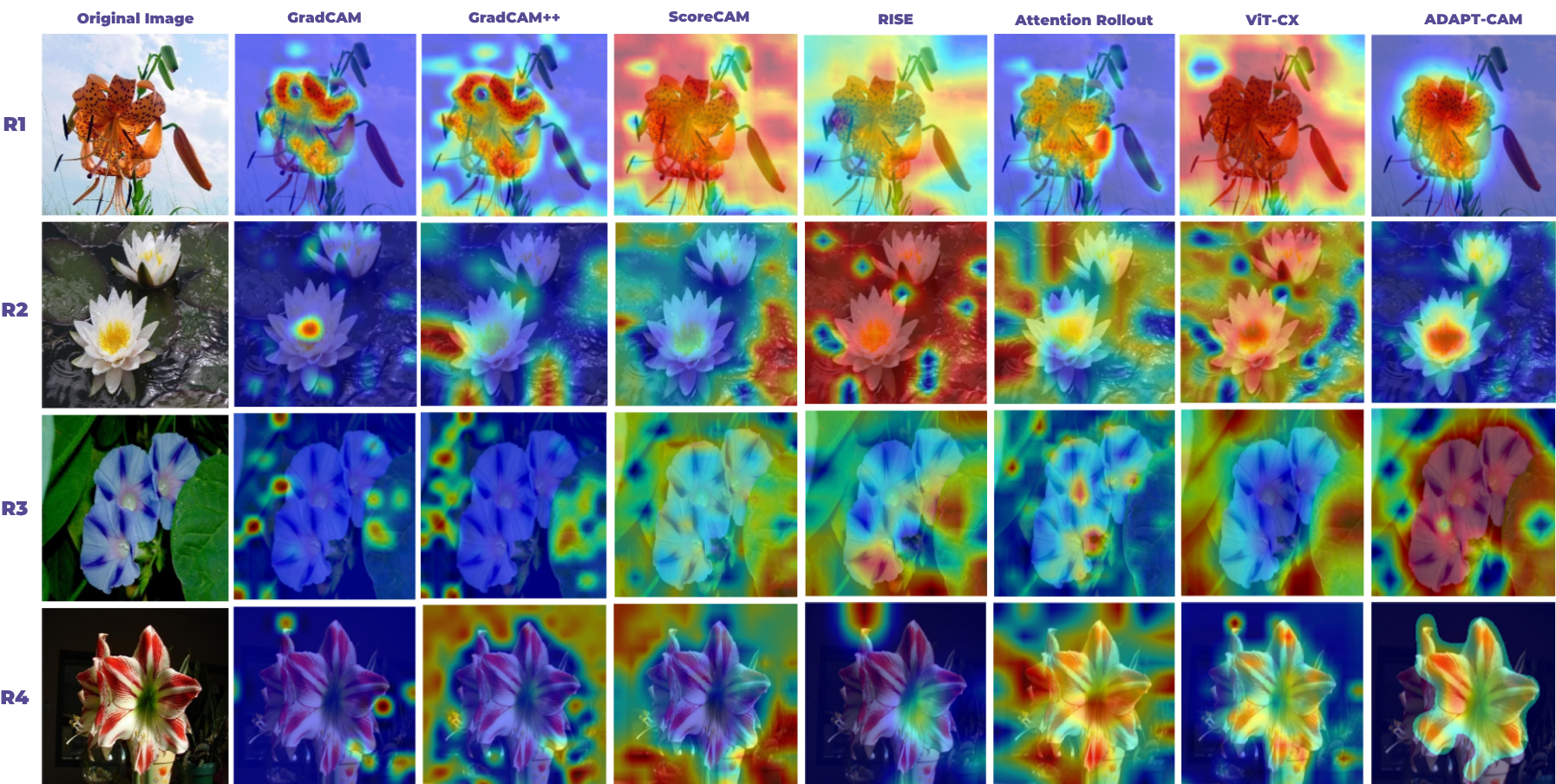}
    \caption{Qualitative comparison on Oxford 102 Flowers~\cite{b32} for ViT-B/16~\cite{b42}}
    \label{figure5}
\end{figure*}

To further evaluate generalization, Fig.~\ref{figure4} and Fig.~\ref{figure5} presents qualitative results on the Oxford 102 Flowers \cite{b32}.

For ResNet-50~\cite{b34} (Fig.~\ref{figure4}), Grad-CAM~\cite{b12} is weak or diffuse (e.g., almost no response in R1, broad background activation in R3). While Grad-CAM++~\cite{b13} and Smooth Grad-CAM++~\cite{b14} cover only parts of the flower and sometimes produce stray blobs (R3, R4). XGrad-CAM~\cite{b18} and Score-CAM~\cite{b15} frequently spill onto leaves or off-object regions (R1, R4). EigenCAM~\cite{b16} is compact on the flower in R1, R2, and R4 but drifts to the background in R3. Only ADAPT-CAM covers the full flower in R1, R2, and R4 and concentrates on the red blooms in R3, with only modest background response.

For ViT-B/16~\cite{b42} (Fig.~\ref{figure5}), Grad-CAM~\cite{b12} and Grad-CAM++~\cite{b13} localize the flower in R1 but become sparse or scattered over the background in R2--R4. Score-CAM~\cite{b15}, RISE~\cite{b39}, Attention Rollout~\cite{b43}, and ViT-CX~\cite{b44} spread substantial activation over the sky, water, and foliage (R1--R3). In R4, the latter two cover the flower together with much of its surroundings. ADAPT-CAM concentrates on the flower in R1, R2, and R4, closely following the petal outline in R4. Whereas in the cluttered R3 scene, its response also extends into the surrounding region.

Overall, ADAPT-CAM yields more compact and object-focused maps than the baselines across both datasets and architectures, though it is not uniformly best in every cluttered scene. Its localization is most consistent on the CNN backbones (DenseNet121 \cite{b33} and ResNet-50 \cite{b34}), with somewhat less consistent behavior on ViT-B/16 \cite{b42} . This is expected, as ADAPT-CAM was originally designed with convolutional architectures in mind. The ViT-B/16 \cite{b42}  experiments serve to test whether the method extends to transformer architectures.

\subsection{Is ADAPT-CAM truly adaptive?}
\label{stability_results}

To further investigate the behaviour of ADAPT-CAM beyond its quantitative and qualitative performance, an additional analysis was conducted to examine the image-adaptive parameterization across different architectures on 200 randomly selected images from ImageNet \cite{b31}. For every image, ADAPT-CAM estimated the power parameter ($p$), the relative curvature parameter ($\lambda_{\mathrm{rel}}$), the number of selected channels, and the size of the adaptive LCN window. Table~\ref{tbl:adaptive_parameters} reports the mean $\pm$ standard deviation of these quantities. The channel counts and window sizes are also expressed as percentages of the available channels and the feature-grid dimension, respectively.

\begin{table*}[t]
\centering
\caption{Cross-architecture comparison of image-adaptive parameters in ADAPT-CAM (mean $\pm$ SD over 200 randomly selected ImageNet images)}
\label{tbl:adaptive_parameters}
\begin{tabular}{lcccc}
\toprule
Parameter & DenseNet-121 \cite{b33} & ResNet-50 \cite{b34} & ViT-B/16 \cite{b42} & DINOv2 \cite{b41} \\
\midrule
$p$ & 0.7848 $\pm$ 0.0068 & 0.6777 $\pm$ 0.0889 & 0.3295 $\pm$ 0.0315 & 0.3998 $\pm$ 0.0186 \\
$\lambda_{\mathrm{rel}}$ & 0.1300 $\pm$ 0.4486 & 0.0018 $\pm$ 0.0052 & 0.0028 $\pm$ 0.0027 & 0.0417 $\pm$ 0.0244 \\
Detected Knee Rank ($K$) & 57.66 $\pm$ 12.12 & 218.91 $\pm$ 75.58 & 316.43 $\pm$ 35.23 & 232.62 $\pm$ 32.13 \\
Detected Knee Rank (\%) & 5.63 $\pm$ 1.18 & 10.69 $\pm$ 3.69 & 41.20 $\pm$ 4.59 & 30.29 $\pm$ 4.18 \\
Adaptive LCN Window & 6.10 $\pm$ 1.00 & 5.07 $\pm$ 0.84 & 9.10 $\pm$ 1.96 & 8.96 $\pm$ 1.90 \\
Adaptive LCN Window (\%) & 87.14 $\pm$ 14.25 & 72.43 $\pm$ 11.94 & 65.00 $\pm$ 14.01 & 56.00 $\pm$ 11.89 \\
\bottomrule
\end{tabular}
\end{table*}

As shown in Table~\ref{tbl:adaptive_parameters}, ADAPT-CAM does not settle on a fixed configuration. The parameters it estimates depend on the feature representation it is applied to. The target layers differ in channel dimensionality and spatial resolution: DenseNet-121 \cite{b33} and ResNet-50 \cite{b34} provide 1024 and 2048 channels on $7\times7$ maps, whereas ViT-B/16 \cite{b42} and DINOv2 \cite{b41} provide 768-dimensional features on $14\times14$ and $16\times16$ patch grids. Raw counts and window sizes are therefore not directly comparable across models, which is why the percentage columns are the meaningful basis for comparison. 

In channel selection, the adaptive knee K, which sets the decay scale of the soft channel mask corresponds to $5.63\%$ and $10.69\%$ of the available channels for DenseNet-121 \cite{b33} and ResNet-50 \cite{b34} as compared with $41.20\%$ and $30.29\%$ for ViT-B/16 \cite{b42} and DINOv2 \cite{b41}. These differences indicate that the transition from dominant to relatively weaker channel contributions occurs at considerably earlier ranks in the convolutional features than in the transformer features. Although ViT-B/16 \cite{b42} has the largest absolute knee rank (316.43), this represents a smaller proportion of its available channels than the corresponding absolute rank might suggest. The detected knee determines the decay scale of the exponential selection mechanism rather than imposing a hard selection boundary. Consequently, smaller knee values produce faster attenuation of lower-ranked channels, whereas larger values result in a more gradual reduction in their contributions. Thus, ADAPT-CAM applies stronger rank-based suppression to the convolutional features and more gradual attenuation to the transformer features. A plausible explanation is that class-relevant contributions are more concentrated among a smaller number of channels in the convolutional representations, whereas they are distributed across a broader set of feature dimensions in the transformer representations. However, this interpretation remains a hypothesis, as channel contribution concentration was not independently investigated. The standard deviations further reveal differences in the image-wise variability of the detected knee. ViT-B/16 \cite{b42} and DINOv2 \cite{b41} exhibit relatively small variation in their knee ranks, whereas ResNet-50 \cite{b34} shows substantially greater variation ($75.58$ channels). This suggests that the contribution transition identified by the selection mechanism is more consistent across images for the transformer models, while varying more considerably with image content for ResNet-50 \cite{b34}.

The estimated power parameter follows the same split. ADAPT-CAM assigns $p$ above 0.67 for the two CNNs, DenseNet-121 \cite{b33} and ResNet-50 \cite{b34}, and around 0.33 to 0.40 for the two transformers, ViT-B/16 \cite{b42} and DINOv2 \cite{b41}. Since $p$ controls how strongly the strongest responses are emphasized, this means ADAPT-CAM applies sharper weighting to the convolutional feature maps and gentler weighting to the patch-based transformer features, where responses are more evenly distributed. The very small spread for DenseNet-121 \cite{b33} ($0.0068$) shows that ADAPT-CAM converges to almost the same $p$ on every image, which may indicate that the estimate sits near the boundary of its allowed range, while the larger spread for ResNet-50 \cite{b34} ($0.0889$) shows more image dependence. The relative curvature term is small on average for ResNet-50 \cite{b34}, ViT-B/16 \cite{b42}, and DINOv2 \cite{b41}, so ADAPT-CAM applies only a minor curvature correction there. For DenseNet-121 \cite{b33} the mean is larger but its standard deviation ($0.4486$) is more than three times the mean, so the distribution is heavily skewed and likely driven by a small number of images with large values. We therefore do not draw strong conclusions from this mean, and a median would summarize it more reliably.

For the LCN window, ADAPT-CAM uses windows covering $87.14\%$ and $72.43\%$ of the feature grid for DenseNet-121 \cite{b33} and ResNet-50 \cite{b34}, and $65.00\%$ and $56.00\%$ for ViT-B/16 \cite{b42} and DINOv2 \cite{b41}. The transformer windows are larger in absolute units (about nine positions against five to six) but smaller relative to their finer grids. Because the CNN maps are only $7\times7$, a window of five or six positions already spans most of the image, so ADAPT-CAM normalizes contrast over broad regions there, while on the finer $14\times14$ and $16\times16$ grids it normalizes over a smaller portion of the image. This shows that ADAPT-CAM scales its spatial context with feature resolution rather than using a fixed window. These patterns are also consistent with the characteristics of ImageNet \cite{b31}, where images typically contain one dominant, often centered object from one of 1000 categories. Such compact, object-centric content makes a small set of channels and a single contiguous spatial region sufficient to explain a prediction, which is compatible with the low retained-channel percentages and large relative windows ADAPT-CAM selects for the CNNs. Since all four models, DenseNet-121 \cite{b33}, ResNet-50 \cite{b34}, ViT-B/16 \cite{b42}, and DINOv2 \cite{b41}, were evaluated on the same 200 images, the differences are unlikely to come from the data, but training regime and pretraining data differ across the models and are confounded with architecture, so we do not attribute them to a single cause. Overall, the parameters estimated by ADAPT-CAM vary with channel dimensionality, grid resolution, and representation type while staying fairly consistent within each model. This supports estimating them per image and per model rather than fixing a single configuration.

\subsection{Does Adaptivity Help?}\label{ablation}

To quantify the contribution of each component of ADAPT-CAM, we conducted an ablation study on DenseNet-121 using 250 images from the ImageNet dataset. Table~\ref{tbl:ablation_study} reports Deletion AUC (lower is better), Insertion AUC (higher is better), Average Drop (lower is better), and Average Increase (higher is better) for ten variants, each obtained by modifying or removing a single component while keeping the rest of the pipeline unchanged. The full ADAPT-CAM configuration achieves the best result on all four metrics (Deletion AUC of 0.1060, Insertion AUC of 0.6738, Average Drop of 10.92\%, and Average Increase of 42.46\%). This indicates that the components are complementary rather than redundant.

\begin{table*}[t]
\centering
\caption{Ablation study of ADAPT-CAM components and adaptive parameter settings}
\label{tbl:ablation_study}
\begin{tabular}{lcccc}
\toprule
Configuration & Deletion AUC & Insertion AUC & Average Drop (\%) & Average Increase (\%) \\
\midrule
A1: No $\lambda_{\mathrm{rel}}$ (Without Curvature) & 0.1617 & 0.6244 & 17.48 & 34.16 \\

A2: $\lambda_{\mathrm{rel}} = 1$ & 0.1280 & 0.6511 & 15.40 & 37.22 \\

A3: $p = 1$ & 0.1253 & 0.6637 & 12.57 & 39.91 \\
A4: $p = 0.1$ & 0.1420 & 0.6448 & 14.25 & 37.62 \\

A5: No Channel Selection & 0.1531 & 0.6356 & 16.16 & 35.50 \\
A6: Hard Channel Selection & 0.1338 & 0.6454 & 13.19 & 38.42 \\

A7: No LCN & 0.1460 & 0.6439 & 18.46 & 35.20 \\
A8: Fixed LCN Window ($=3$) & 0.1243 & 0.6645 & 16.34 & 36.65 \\
A9: Fixed LCN Window ($=7$) & 0.1326 & 0.6558 & 15.64 & 38.18 \\

A10: No Gaussian Smoothing & 0.1169 & 0.6650 & 13.72 & 40.78 \\
\textbf{ADAPT-CAM (Baseline)} & \textbf{0.1060} & \textbf{0.6738} & \textbf{10.92} & \textbf{42.46} \\
\bottomrule
\end{tabular}
\end{table*}

Removing the curvature term (A1, no $\lambda_{\mathrm{rel}}$) causes the largest degradation in Deletion AUC (0.1617) and Insertion AUC (0.6244), showing that second-order information is the most influential component. Replacing the adaptive $\lambda_{\mathrm{rel}}$ with a fixed value of 1 (A2) also hurts performance. This suggests that the curvature weighting must be calibrated per image. Likewise, fixing the exponent to $p=1$ (A3) or $p=0.1$ (A4) yields inferior results. This shows the benefit of deriving $p$ from the energy decay of the channels rather than fixing it. Disabling channel selection (A5) degrades all metrics, and replacing the knee-guided soft selection with a hard cutoff (A6) is also suboptimal. Removing LCN (A7) leads to the highest Average Drop (18.46\%). Even fixed LCN window sizes of 3 and 7 (A8, A9) both underperform the adaptive kernel. This suggests that the window size should follow the spatial extent of the salient activations. Finally, omitting Gaussian smoothing (A10) gives a smaller but consistent loss, indicating that it helps suppress residual noise in the saliency map.

\section{Conclusion}
\label{conclusion}

This study proposed ADAPT-CAM, a novel gradient-based, parameter-adaptive CAM framework inspired by contextual contrast processing and selective response modulation in biological vision. It combines first-order gradient energy with an activation-weighted second-order curvature term to produce power-scaled channel weighting. It further integrates adaptive LCN whose neighborhood size is determined adaptively through a participation-ratio-based mechanism. An adaptive soft channel selection stage based on cumulative contribution knee-point detection and a knee-guided exponential rank-based mask is used to progressively attenuate lower-ranked channels without completely discarding them. Most importantly, the key parameters governing these operations are estimated independently for each input rather than fixed across the dataset or architecture. This enables ADAPT-CAM to highlight class-relevant regions while reducing the influence of weak and noisy channels, producing precise, stable and class-specific activation maps. Experiments on ImageNet \cite{b31} and Oxford 102 Flowers \cite{b32} using CNN and Transformer architectures show that ADAPT-CAM outperforms established CAM baselines in Insertion AUC, Deletion AUC, Average Drop and Average Increase. Qualitative evaluations further illustrate its ability to highlight relevant image regions while preserving spatial detail and attenuating less informative contributions. These results support the robustness and generalizability of the approach across datasets, architectures and task granularities. By providing more faithful and reliable explanations, ADAPT-CAM represents a meaningful advancement in gradient-based XAI.Future work will investigate the applicability of ADAPT-CAM to additional domains, particularly medical imaging, and explore its extension to diverse model architectures and tasks.


\printcredits

\section*{Declaration of Generative AI and AI-Assisted Technologies in the Manuscript Preparation Process}

During the preparation of this work, the authors used OpenAI's ChatGPT solely to improve grammatical syntax and sentence cohesion. All content was subsequently reviewed and verified by the authors. No AI technology was used as a substitute for human critical thinking, engineering evaluation, or original data interpretation. The authors take full responsibility for the scientific accuracy and integrity of this article.


\bibliographystyle{unsrtnat}
\bibliography{cas_refs}

@article{b4,
title = {A review of convolutional neural network based methods for medical image classification},
journal = {Computers in Biology and Medicine},
volume = {185},
pages = {109507},
year = {2025},
issn = {0010-4825},
doi = {https://doi.org/10.1016/j.compbiomed.2024.109507},
url = {https://www.sciencedirect.com/science/article/pii/S0010482524015920},
author = {Chao Chen and Nor Ashidi {Mat Isa} and Xin Liu}
}

@ARTICLE{b5,
  author={Naseer, Aysha and Mudawi, Naif Al and Abdelhaq, Maha and Alonazi, Mohammed and Alazeb, Abdulwahab and Algarni, Asaad and Jalal, Ahmad},
  journal={IEEE Access}, 
  title={CNN-Based Object Detection via Segmentation Capabilities in Outdoor Natural Scenes}, 
  year={2024},
  volume={12},
  number={},
  pages={84984-85000},
  doi={10.1109/ACCESS.2024.3413848}}

@article{b6,
  author  = {Muzammul, Muhamad and Li, Xi},
  title   = {Comprehensive review of deep learning-based tiny object detection: challenges, strategies, and future directions},
  journal = {Knowledge and Information Systems},
  volume  = {67},
  number  = {5},
  pages   = {3825--3913},
  year    = {2025},
  doi     = {10.1007/s10115-025-02375-9},
  issn    = {0219-3116},
  url     = {https://doi.org/10.1007/s10115-025-02375-9}
}

@article{b7,
  author  = {Koishiyeva, D. and Mukhammejanova, D. and Kang, J. W. and Mukasheva, A.},
  title   = {A Review of Deep Learning Approaches Based on Segment Anything Model for Medical Image Segmentation},
  journal = {Bioengineering},
  volume  = {12},
  year    = {2025},
  pages   = {1312},
  doi     = {10.3390/bioengineering12121312},
  url     = {https://doi.org/10.3390/bioengineering12121312}
}

@article{b8,
title = {A comprehensive review of deep learning for medical image segmentation},
journal = {Neurocomputing},
volume = {613},
pages = {128740},
year = {2025},
issn = {0925-2312},
doi = {https://doi.org/10.1016/j.neucom.2024.128740},
url = {https://www.sciencedirect.com/science/article/pii/S092523122401511X},
author = {Qingling Xia and Hong Zheng and Haonan Zou and Dinghao Luo and Hongan Tang and Lingxiao Li and Bin Jiang}
}

@InProceedings{b9,
    author    = {Du, Ruoyi and Yu, Wenqing and Wang, Heqing and Lin, Ting-En and Chang, Dongliang and Ma, Zhanyu},
    title     = {Multi-View Active Fine-Grained Visual Recognition},
    booktitle = {Proceedings of the IEEE/CVF International Conference on Computer Vision (ICCV)},
    month     = {October},
    year      = {2023},
    pages     = {1568-1578}
}

@INPROCEEDINGS{b11,
  author={Zhou, Bolei and Khosla, Aditya and Lapedriza, Agata and Oliva, Aude and Torralba, Antonio},
  booktitle={2016 IEEE Conference on Computer Vision and Pattern Recognition (CVPR)}, 
  title={Learning Deep Features for Discriminative Localization}, 
  year={2016},
  volume={},
  number={},
  pages={2921-2929},
  doi={10.1109/CVPR.2016.319}}

@INPROCEEDINGS{b12,
  author={Selvaraju, Ramprasaath R. and Cogswell, Michael and Das, Abhishek and Vedantam, Ramakrishna and Parikh, Devi and Batra, Dhruv},
  booktitle={2017 IEEE International Conference on Computer Vision (ICCV)}, 
  title={Grad-CAM: Visual Explanations from Deep Networks via Gradient-Based Localization}, 
  year={2017},
  volume={},
  number={},
  pages={618-626},
}

@INPROCEEDINGS{b13,
  author={Chattopadhay, Aditya and Sarkar, Anirban and Howlader, Prantik and Balasubramanian, Vineeth N},
  booktitle={2018 IEEE Winter Conference on Applications of Computer Vision (WACV)}, 
  title={Grad-CAM++: Generalized Gradient-Based Visual Explanations for Deep Convolutional Networks}, 
  year={2018},
  volume={},
  number={},
  pages={839-847},
  doi={10.1109/WACV.2018.00097}}

@article{b14,
  title={Smooth Grad-CAM++: An Enhanced Inference Level Visualization Technique for Deep Convolutional Neural Network Models},
  author={Daniel Omeiza and Skyler Speakman and Celia Cintas and Komminist Weldemariam},
  journal={ArXiv},
  year={2019},
  volume={abs/1908.01224},
  url={https://api.semanticscholar.org/CorpusID:199442547}
}

@INPROCEEDINGS {b15,
author = { Wang, Haofan and Wang, Zifan and Du, Mengnan and Yang, Fan and Zhang, Zijian and Ding, Sirui and Mardziel, Piotr and Hu, Xia },
booktitle = { 2020 IEEE/CVF Conference on Computer Vision and Pattern Recognition Workshops (CVPRW) },
title = {{ Score-CAM: Score-Weighted Visual Explanations for Convolutional Neural Networks }},
year = {2020},
volume = {},
ISSN = {},
pages = {111-119},
doi = {10.1109/CVPRW50498.2020.00020},
url = {https://doi.ieeecomputersociety.org/10.1109/CVPRW50498.2020.00020},
publisher = {IEEE Computer Society},
address = {Los Alamitos, CA, USA},
month =Jun}

@article{b16,
  author  = {Bany Muhammad, Mohammed and Yeasin, Mohammed},
  title   = {Eigen-CAM: Visual Explanations for Deep Convolutional Neural Networks},
  journal = {SN Computer Science},
  volume  = {2},
  number  = {1},
  pages   = {47},
  year    = {2021},
  doi     = {10.1007/s42979-021-00449-3},
  url     = {https://doi.org/10.1007/s42979-021-00449-3},
  issn    = {2661-8907}
}

@article{b17,
author = {Jiang, Peng-Tao and Zhang, Chang-Bin and Hou, Qibin and Cheng, Ming-Ming and Wei, Yunchao},
title = {LayerCAM: Exploring Hierarchical Class Activation Maps for Localization},
year = {2021},
issue_date = {2021},
publisher = {IEEE Press},
volume = {30},
issn = {1057-7149},
url = {https://doi.org/10.1109/TIP.2021.3089943},
doi = {10.1109/TIP.2021.3089943},
journal = {Trans. Img. Proc.},
month = jan,
pages = {5875–5888},
numpages = {14}
}

@misc{b18,
      title={Axiom-based Grad-CAM: Towards Accurate Visualization and Explanation of CNNs}, 
      author={Ruigang Fu and Qingyong Hu and Xiaohu Dong and Yulan Guo and Yinghui Gao and Biao Li},
      year={2020},
      eprint={2008.02312},
      archivePrefix={arXiv},
      primaryClass={cs.CV},
      url={https://arxiv.org/abs/2008.02312}, 
}

@article{b19,
title = {Master-CAM: Multi-scale fusion guided by Master map for high-quality class activation maps},
journal = {Displays},
volume = {76},
pages = {102339},
year = {2023},
issn = {0141-9382},
doi = {https://doi.org/10.1016/j.displa.2022.102339},
url = {https://www.sciencedirect.com/science/article/pii/S0141938222001573},
author = {Xuesheng Zhou and Yan Li and Guitao Cao and Wenming Cao},
}

@INPROCEEDINGS{b20,
  author={Sritharan, Niruthikka and Gnanavel, Nishaanthini and Inparaj, Prathushan and Meedeniya, Dulani and Yogarajah, Pratheepan},
  booktitle={2024 International Research Conference on Smart Computing and Systems Engineering (SCSE)}, 
  title={EnsembleCAM: Unified Visualization for Explainable Cervical Cancer Identification}, 
  year={2024},
  volume={7},
  number={},
  pages={1-6},
  doi={10.1109/SCSE61872.2024.10550859}}

@ARTICLE{b21,
  author={Soomro, Shafiullah and Niaz, Asim and Nam Choi, Kwang},
  journal={IEEE Access}, 
  title={Grad++ScoreCAM: Enhancing Visual Explanations of Deep Convolutional Networks Using Incremented Gradient and Score- Weighted Methods}, 
  year={2024},
  volume={12},
  number={},
  pages={61104-61112},
  doi={10.1109/ACCESS.2024.3392853}}

@article{b22,
  author  = {Cheng, Zhihan and Wu, Yue and Li, Yule and Cai, Lingfeng and Ihnaini, Baha},
  title   = {A Comprehensive Review of Explainable Artificial Intelligence (XAI) in Computer Vision},
  journal = {Sensors},
  year    = {2025},
  volume  = {25},
  number  = {13},
  pages   = {4166},
  doi     = {10.3390/s25134166},
  url     = {https://doi.org/10.3390/s25134166}
}

@Article{b23,
AUTHOR = {Gao, Yongshun and Liu, Jie and Li, Weihan and Hou, Ming and Li, Yang and Zhao, Huimin},
TITLE = {Augmented Grad-CAM++: Super-Resolution Saliency Maps for Visual Interpretation of Deep Neural Network},
JOURNAL = {Electronics},
VOLUME = {12},
YEAR = {2023},
NUMBER = {23},
ARTICLE-NUMBER = {4846},
URL = {https://www.mdpi.com/2079-9292/12/23/4846},
ISSN = {2079-9292},
DOI = {10.3390/electronics12234846}
}

@article{b24,
title = {Artificial intelligence in healthcare delivery: Prospects and pitfalls},
journal = {Journal of Medicine, Surgery, and Public Health},
volume = {3},
pages = {100108},
year = {2024},
issn = {2949-916X},
doi = {https://doi.org/10.1016/j.glmedi.2024.100108},
url = {https://www.sciencedirect.com/science/article/pii/S2949916X24000616},
author = {David B. Olawade and Aanuoluwapo C. David-Olawade and Ojima Z. Wada and Akinsola J. Asaolu and Temitope Adereni and Jonathan Ling}
}

@article{b25,
author = {Zhao, Jingyuan and Wu, Yuyan and Deng, Rui and Xu, Susu and Gao, Jinpeng and Burke, Andrew},
title = {A Survey of Autonomous Driving from a Deep Learning Perspective},
year = {2025},
issue_date = {October 2025},
publisher = {Association for Computing Machinery},
address = {New York, NY, USA},
volume = {57},
number = {10},
issn = {0360-0300},
url = {https://doi.org/10.1145/3729420},
doi = {10.1145/3729420},
month = may,
articleno = {263},
numpages = {60}
}

@article{b26,
title = {Explainable Artificial Intelligence (XAI): What we know and what is left to attain Trustworthy Artificial Intelligence},
journal = {Information Fusion},
volume = {99},
pages = {101805},
year = {2023},
issn = {1566-2535},
doi = {https://doi.org/10.1016/j.inffus.2023.101805},
url = {https://www.sciencedirect.com/science/article/pii/S1566253523001148},
author = {Sajid Ali and Tamer Abuhmed and Shaker El-Sappagh and Khan Muhammad and Jose M. Alonso-Moral and Roberto Confalonieri and Riccardo Guidotti and Javier {Del Ser} and Natalia Díaz-Rodríguez and Francisco Herrera}
}

@article{b27,
  author  = {Linardatos, Pantelis and Papastefanopoulos, Vasilis and Kotsiantis, Sotiris},
  title   = {Explainable AI: A Review of Machine Learning Interpretability Methods},
  journal = {Entropy},
  volume  = {23},
  number  = {1},
  pages   = {18},
  year    = {2020},
  doi     = {10.3390/e23010018},
  issn    = {1099-4300},
  url     = {https://doi.org/10.3390/e23010018},
  pmid    = {33375658},
  pmcid   = {PMC7824368}
}

@ARTICLE{b28,
AUTHOR={Cheong, Ben Chester },    
TITLE={Transparency and accountability in AI systems: safeguarding wellbeing in the age of algorithmic decision-making},  
JOURNAL={Frontiers in Human Dynamics},
VOLUME={Volume 6 - 2024},
YEAR={2024},
URL={https://www.frontiersin.org/journals/human-dynamics/articles/10.3389/fhumd.2024.1421273},
DOI={10.3389/fhumd.2024.1421273},
ISSN={2673-2726},
}

@article{b29,
author = {Perez-Cerrolaza, Jon and Abella, Jaume and Borg, Markus and Donzella, Carlo and Cerquides, Jes\'{u}s and Cazorla, Francisco J. and Englund, Cristofer and Tauber, Markus and Nikolakopoulos, George and Flores, Jose Luis},
title = {Artificial Intelligence for Safety-Critical Systems in Industrial and Transportation Domains: A Survey},
year = {2024},
issue_date = {July 2024},
publisher = {Association for Computing Machinery},
address = {New York, NY, USA},
volume = {56},
number = {7},
issn = {0360-0300},
url = {https://doi.org/10.1145/3626314},
doi = {10.1145/3626314},
journal = {ACM Comput. Surv.},
month = apr,
articleno = {176},
numpages = {40},
}

@article{b30,
  author  = {Rai, Arun},
  title   = {Explainable AI: from black box to glass box},
  journal = {Journal of the Academy of Marketing Science},
  volume  = {48},
  number  = {1},
  pages   = {137--141},
  year    = {2020},
  doi     = {10.1007/s11747-019-00710-5},
  issn    = {1552-7824},
  url     = {https://doi.org/10.1007/s11747-019-00710-5}
}

@INPROCEEDINGS{b31,
  author={Deng, Jia and Dong, Wei and Socher, Richard and Li, Li-Jia and Kai Li and Li Fei-Fei},
  booktitle={2009 IEEE Conference on Computer Vision and Pattern Recognition}, 
  title={ImageNet: A large-scale hierarchical image database}, 
  year={2009},
  volume={},
  number={},
  pages={248-255},
  doi={10.1109/CVPR.2009.5206848}}

@InProceedings{b32,
  author    = "Maria-Elena Nilsback and Andrew Zisserman",
  title     = "Automated Flower Classification over a Large Number of Classes",
  booktitle = "Proceedings of the Indian Conference on Computer Vision, Graphics and Image Processing",
  month     = "Dec",
  year      = "2008"
}

@INPROCEEDINGS{b33,
  author={Huang, Gao and Liu, Zhuang and Van Der Maaten, Laurens and Weinberger, Kilian Q.},
  booktitle={2017 IEEE Conference on Computer Vision and Pattern Recognition (CVPR)}, 
  title={Densely Connected Convolutional Networks}, 
  year={2017},
  volume={},
  number={},
  pages={2261-2269},
  doi={10.1109/CVPR.2017.243}}

@INPROCEEDINGS{b34,
  author={He, Kaiming and Zhang, Xiangyu and Ren, Shaoqing and Sun, Jian},
  booktitle={2016 IEEE Conference on Computer Vision and Pattern Recognition (CVPR)}, 
  title={Deep Residual Learning for Image Recognition}, 
  year={2016},
  volume={},
  number={},
  pages={770-778},
  doi={10.1109/CVPR.2016.90}}

@book{b37,
  title={Digital Image Processing},
  author={Gonzalez, Rafael C and Woods, Richard E},
  edition={4th},
  year={2018},
  publisher={Pearson}
}

@inproceedings{b39,
  title={RISE: Randomized Input Sampling for Explanation of Black-box Models},
  author={Petsiuk, Vladimir and Das, Abir and Saenko, Kate},
  booktitle={British Machine Vision Conference (BMVC)},
  year={2018}
}

@article{b40,
  author  = {Chen, Zhiqing and Zhang, Yuejin J. and Pan, Lei and Lei, Zhi and Zhou, Zhiyuan and Huang, Jinsuo S.},
  title   = {{SSG--CAM}: Enhancing Visual Interpretability through Refined Second-Order Gradients and Evolutionary Multi-Layer Fusion},
  journal = {Scientific Reports},
  year    = {2026},
  volume  = {16},
  number  = {1},
  pages   = {6848},
  doi     = {10.1038/s41598-026-37278-4},
  url     = {https://doi.org/10.1038/s41598-026-37278-4}
}

@article{
b41,
title={{DINO}v2: Learning Robust Visual Features without Supervision},
author={Maxime Oquab and Timoth{\'e}e Darcet and Th{\'e}o Moutakanni and Huy V. Vo and Marc Szafraniec and Vasil Khalidov and Pierre Fernandez and Daniel HAZIZA and Francisco Massa and Alaaeldin El-Nouby and Mido Assran and Nicolas Ballas and Wojciech Galuba and Russell Howes and Po-Yao Huang and Shang-Wen Li and Ishan Misra and Michael Rabbat and Vasu Sharma and Gabriel Synnaeve and Hu Xu and Herve Jegou and Julien Mairal and Patrick Labatut and Armand Joulin and Piotr Bojanowski},
journal={Transactions on Machine Learning Research},
issn={2835-8856},
year={2024},
url={https://openreview.net/forum?id=a68SUt6zFt},
note={Featured Certification}
}

@inproceedings{
b42,
title={An Image is Worth 16x16 Words: Transformers for Image Recognition at Scale},
author={Alexey Dosovitskiy and Lucas Beyer and Alexander Kolesnikov and Dirk Weissenborn and Xiaohua Zhai and Thomas Unterthiner and Mostafa Dehghani and Matthias Minderer and Georg Heigold and Sylvain Gelly and Jakob Uszkoreit and Neil Houlsby},
booktitle={International Conference on Learning Representations},
year={2021},
url={https://openreview.net/forum?id=YicbFdNTTy}
}

@inproceedings{b43,
  title={Quantifying Attention Flow in Transformers},
  author={Abnar, Samira and Zuidema, Willem},
  booktitle={Proceedings of the 58th Annual Meeting of the Association for Computational Linguistics (ACL)},
  pages={4190--4197},
  year={2020},
  address={Virtual Event},
  publisher={Association for Computational Linguistics}
}

@inproceedings{b44,
author = {Xie, Weiyan and Li, Xiao-Hui and Cao, Caleb Chen and Zhang, Nevin L.},
title = {ViT-CX: causal explanation of vision transformers},
year = {2023},
isbn = {978-1-956792-03-4},
booktitle = {Proceedings of the Thirty-Second International Joint Conference on Artificial Intelligence},
articleno = {174},
numpages = {9},
location = {Macao, P.R.China},
series = {IJCAI '23}
}

@article{b45,
  author    = {Carandini, Matteo and Heeger, David J.},
  title     = {Normalization as a Canonical Neural Computation},
  journal   = {Nature Reviews Neuroscience},
  year      = {2012},
  volume    = {13},
  number    = {1},
  pages     = {51--62},
  doi       = {10.1038/nrn3136},
  url       = {https://doi.org/10.1038/nrn3136}
}

@article{b46,
  author    = {Reynolds, John H. and Heeger, David J.},
  title     = {The Normalization Model of Attention},
  journal   = {Neuron},
  year      = {2009},
  volume    = {61},
  number    = {2},
  pages     = {168--185},
  doi       = {10.1016/j.neuron.2009.01.002},
  url       = {https://doi.org/10.1016/j.neuron.2009.01.002},
  pmid      = {19186161},
  pmcid     = {PMC2752446}
}



\end{document}